%% file: anonymous-submission-latex-2026.tex
\documentclass[letterpaper]{article} 
\usepackage{aaai2026}  
\usepackage{times}  
\usepackage{helvet}  
\usepackage{courier}  
\usepackage[hyphens]{url}  
\usepackage{graphicx} 
\usepackage{natbib}  
\usepackage{caption} 
\usepackage{algorithm}
\usepackage{algorithmic}

\usepackage{newfloat}
\usepackage{listings}
\DeclareCaptionStyle{ruled}{labelfont=normalfont,labelsep=colon,strut=off} 
\floatstyle{ruled}
\newfloat{listing}{tb}{lst}{}
\floatname{listing}{Listing}
\usepackage[table,dvipsnames]{xcolor}
\usepackage{tikz}
\usepackage[most]{tcolorbox}
\usepackage{adjustbox}
\usepackage{marvosym}
\usepackage{amsfonts}
\usepackage{xspace}
\usepackage{diagbox, makecell}
\usepackage{booktabs}
\usepackage{multirow}
\usepackage{multicol}
\usepackage{newtxtext,newtxmath}
\usepackage{enumitem}
\usepackage{tocloft}

\definecolor{blue1}{HTML}{75B0CF}
\definecolor{blue2}{HTML}{8bb0d0}
\definecolor{blue3}{HTML}{F2F6F8}

\newcommand{\obv}{\textsc{\textbf{Oblivionis}}\@\xspace}
\newcommand{\eg}{e.g.\@\xspace}       
\newcommand{\etal}{et~al.\@\xspace}   
\newcommand{\inlineicon}[2][height=1.2em]{\raisebox{-0.2ex}{\includegraphics[#1]{#2}}}

\newtcolorbox{cmt}[2][]{
  colbacktitle = blue1,
  colframe = blue1,
  colback = blue3,
  coltitle = white,
  fonttitle=\bfseries,
  title={QA: #2},
  #1 
}

\newtcolorbox{pmt}[2][]{
  colbacktitle = blue1,
  colframe = blue1,
  colback = blue3,
  coltitle = white,
  fonttitle=\bfseries,
  title={Prompt: #2},
  #1 
}

\title{\inlineicon{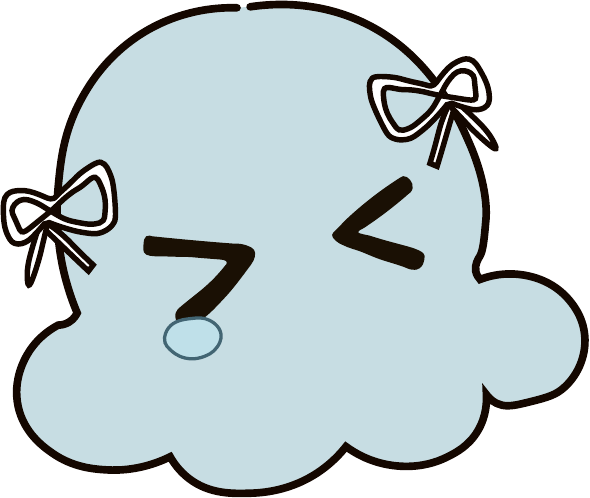}~\obv: A Lightweight Learning and Unlearning Framework for\\Federated Large Language Models}

\author{
    Fuyao Zhang\textsuperscript{$\heartsuit$}\textsuperscript{\equalcontrib}, Xinyu Yan\textsuperscript{$\heartsuit$}\textsuperscript{\equalcontrib}, Tiantong Wu\textsuperscript{$\heartsuit$}, Wenjie Li\textsuperscript{$\heartsuit$,}\textsuperscript{$\triangle$}, Tianxiang Chen\textsuperscript{$\heartsuit$}\\
    Yang Cao\textsuperscript{$\clubsuit$}, Ran Yan\textsuperscript{$\heartsuit$}, Longtao Huang\textsuperscript{$\blacklozenge$}, Wei Yang Bryan Lim\textsuperscript{$\heartsuit$}\textsuperscript{\thanks{Corresponding Author.}}, Qiang Yang\textsuperscript{$\spadesuit$}
}
\affiliations{
    \textsuperscript{$\heartsuit$} Nanyang Technological University \quad \textsuperscript{$\triangle$} Hebei Normal University \quad \\
    \textsuperscript{$\clubsuit$} Institute of Science Tokyo \quad 
    \textsuperscript{$\blacklozenge$} Alibaba Group \quad 
    \textsuperscript{$\spadesuit$} Hong Kong Polytechnic University

}

\usepackage{bibentry}

\begin{document}

\maketitle


\begin{abstract}

Large Language Models (LLMs) increasingly leverage Federated Learning (FL) to utilize private, task-specific datasets for fine-tuning while preserving data privacy. However, while federated LLM frameworks effectively enable collaborative training without raw data sharing, they critically lack built-in mechanisms for regulatory compliance like GDPR’s \textit{right to be forgotten}. Integrating private data heightens concerns over data quality and long-term governance, yet existing distributed training frameworks offer no principled way to selectively remove specific client contributions post-training. Due to distributed data silos, stringent privacy constraints, and the intricacies of interdependent model aggregation, federated LLM unlearning is significantly more complex than centralized LLM unlearning. To address this gap, we introduce \obv, a lightweight learning and unlearning framework that enables clients to selectively remove specific private data during federated LLM training, enhancing trustworthiness and regulatory compliance. By unifying FL and unlearning as a dual optimization objective, we incorporate $6$ FL and $5$ unlearning algorithms for comprehensive evaluation and comparative analysis, establishing a robust pipeline for federated LLM unlearning. Extensive experiments demonstrate that \obv outperforms local training, achieving a robust balance between forgetting efficacy and model utility, with cross-algorithm comparisons providing clear directions for future LLM development.

\end{abstract}


\begin{links}
    \link{Code}{https://github.com/fyzhang1/Oblivionis}

    \link{Project Page}{https://fyzhang1.github.io/Oblivionis}
\end{links}

\input{Section/1-Introduction}

\input{Section/2-RelatedWork}
\input{Section/3-Overview}
\input{Section/4-Experiment}

\input{Section/5-Conclusion}

\bibliography{aaai2026}

\input{Section/Appendix}

\end{document}

%% file: Section/1-Introduction.tex
\section{Introduction}



Large Language Models (LLMs), driven by the Transformer architecture~\cite{vaswani2017attention}, have transformed Natural Language Processing and diverse fields~\cite{achiam2023gpt,touvron2023llama}. By efficiently learning complex patterns from vast datasets, they enable advanced tasks such as text generation, translation, and question-answering~\cite{wei2022chain, webb2023emergent,imani2023mathprompter}.
Typically, the increase in the quantity and quality of data samples leads to stronger generalization capabilities and higher task accuracy. In particular, LLM fine-tuning relies on limited task-specific private data. Such data cannot be used for centralized fine-tuning as it often involves personal information or holds significant economic value, as seen in domains like the medical and financial~\cite{thirunavukarasu2023large, wu2023bloomberggpt}.





\begin{figure}[t]
    \centering
    \includegraphics[width=\linewidth]{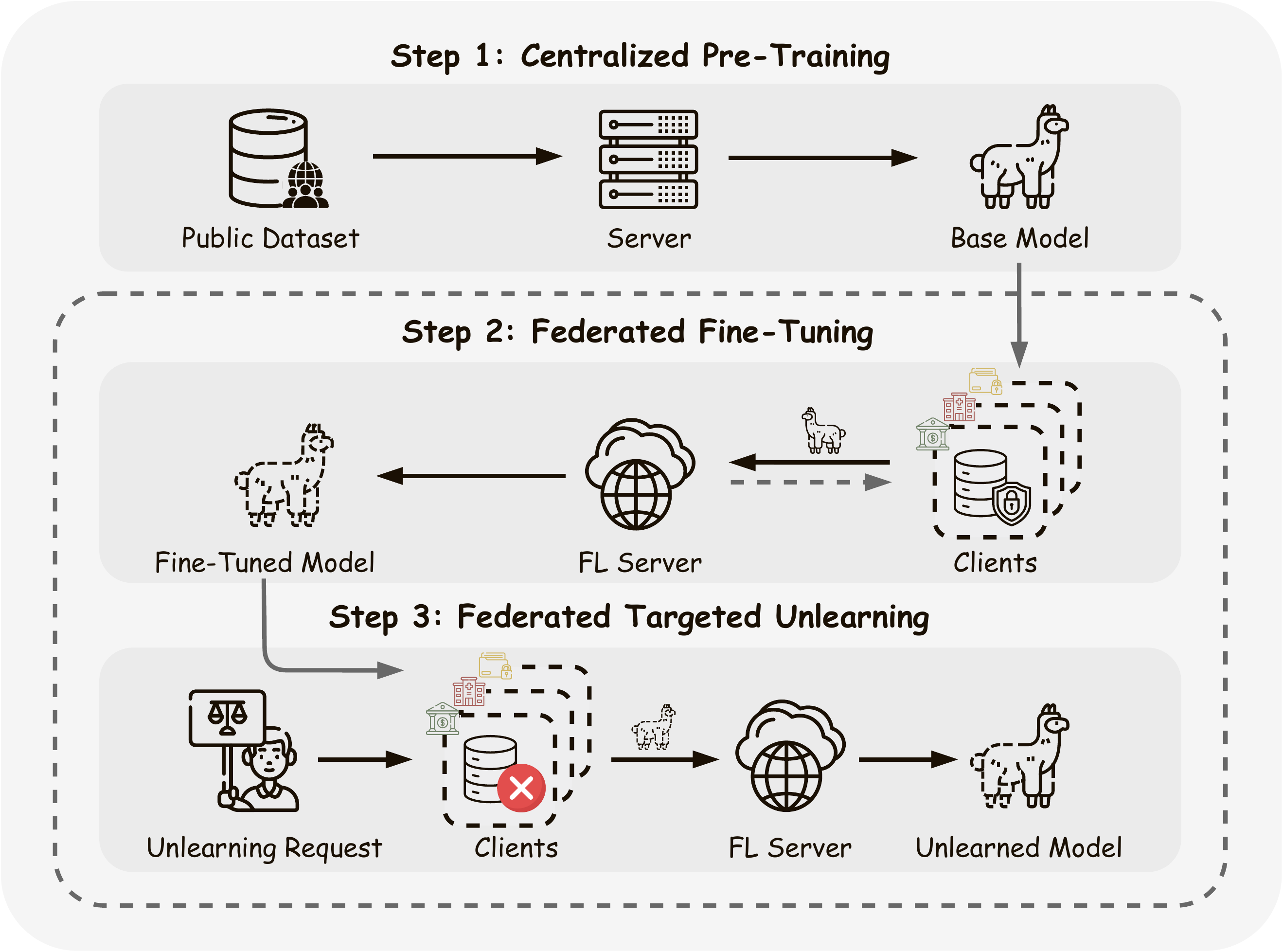}
    \caption{\textbf{Illustration of the three-step LLM training process}: (1) Pre-training the base model with public datasets on a centralized server; (2) Federated fine-tuning on the base model using private and sensitive task-specific data~\inlineicon{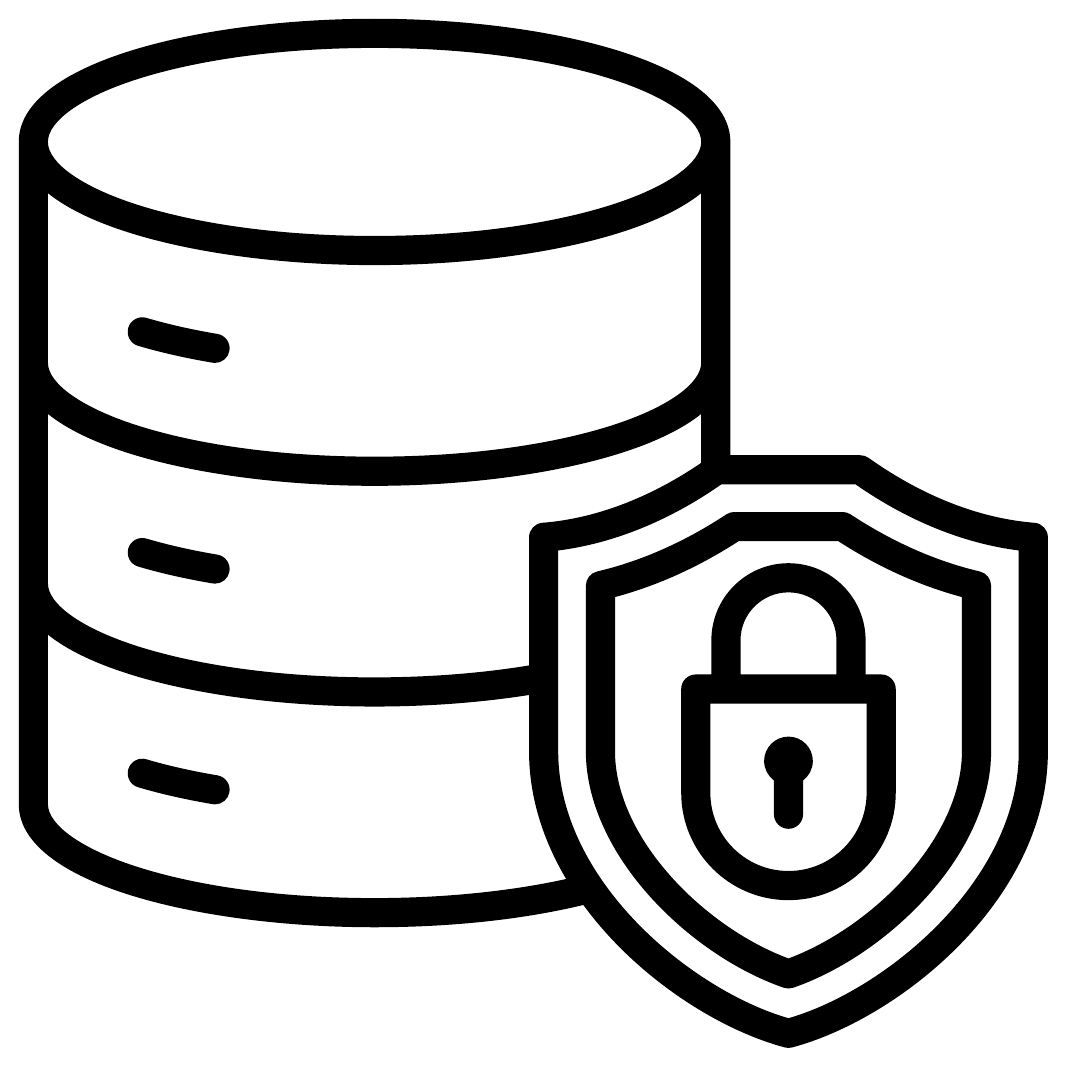}; (3) Federated targeted unlearning removes the influence of specific data~\inlineicon{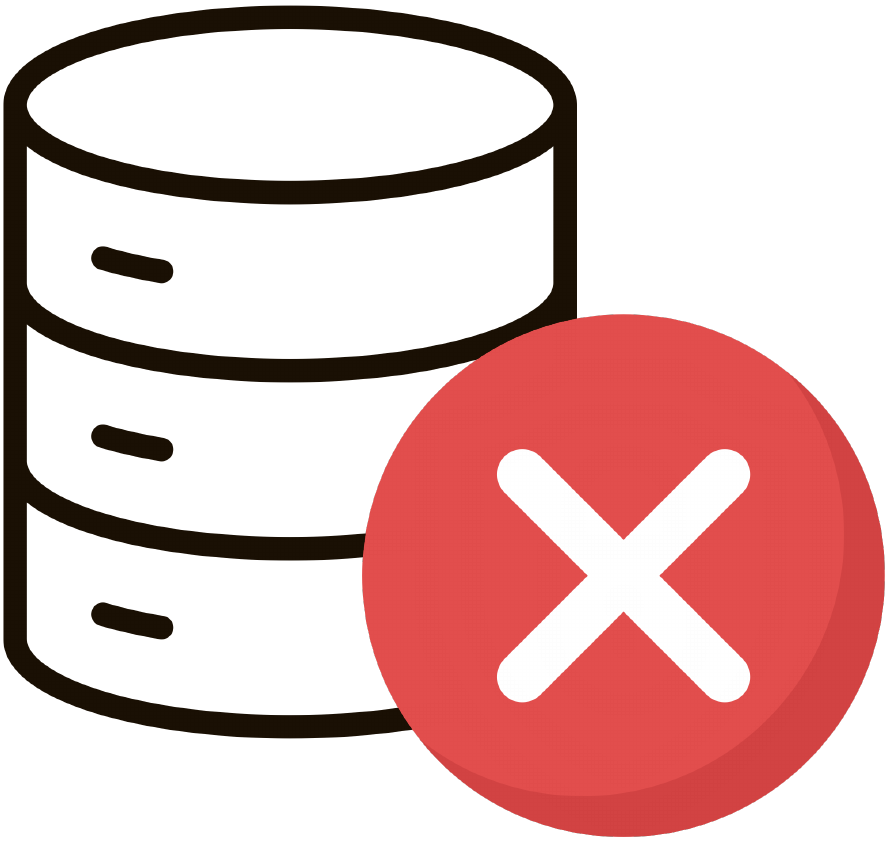} upon client requests, addressing regulatory and ethical requirements. Areas enclosed by grey dashed boxes are our main contributions.}
    \label{fig:step}
\end{figure}

In this context, Federated Learning (FL)~\cite{mcmahan2017communication}, as an emerging distributed machine learning paradigm, becomes a highly anticipated trend in the development of LLM training because of its unique collaborative training mechanism and inherent privacy-preserving feature. Federated LLM (FedLLM) allows multiple clients to jointly fine-tune a global model without sharing their local private data. Specifically, Chen~\etal~(\citeyear{chen2023federated}) first proposed a systematic research framework to explore the integration between LLM and FL. 
Fan~\etal~(\citeyear{fan2023fate}) proposed an industrial-grade framework for FedLLM that addresses resource consumption and data privacy challenges, supporting efficient training and privacy-preserving mechanisms. Ye~\etal~(\citeyear{ye2024openfedllm}) proposed the OpenFedLLM framework for training LLM on decentralized private data, with federated instruction tuning, value alignment, and multiple FL algorithms. 

Although FL offers a promising approach for the continuous evolution of LLMs, it still encounters significant challenges in practical applications. As depicted in Figure~\ref{fig:step}, the large number of participating FL clients and diverse data sources can lead to global models inadvertently learning low-quality knowledge, biased information, or outdated content from specific clients during federated fine-tuning~\cite{wei2023jailbroken, min2023silo}. Furthermore, as global data privacy regulations (\eg, the E.U.'s General Data Protection Regulation, GDPR) become increasingly sophisticated and public awareness of user data rights grows, the right to be forgotten and data deletion requests are gaining more importance~\cite{rosen2011right,pardau2018california}. Thus, LLMs require not only the capability to acquire new knowledge, but also the ability to effectively remove specific data and its contribution to the model upon the removal request~\cite{huu2024effects,wang2024llm,wang2025rethinking}. Preventing model retention of removed data is critical for maintaining user trust, ensuring regulatory compliance, and preserving model integrity.

Based on the above challenges and requirements, we aim to explore an innovative LLM training paradigm to effectively mitigate the influence of low-quality knowledge within the FL framework and empower the model to respond to data contribution removal requests. We propose that during the training process of FedLLM, when a client opts out of FL or its data contribution legally needs to be removed, the global model should be able to perform federated targeted unlearning. This process is designed to achieve three key objectives: \textbf{(1) Effectiveness}, selectively removing all influences of a client’s local private data from the global model; \textbf{(2) Robustness}, ensuring the model maintains high utility on retained data; \textbf{(3) Lightweight Design}, enabling unlearning with minimal computational resources and model parameters. To achieve these goals, we propose \obv, a lightweight FedLLM unlearning framework that integrates federated fine-tuning and targeted unlearning, enabling robust LLM training while ensuring compliance with privacy regulations. In conclusion, our contributions are as follows:

\begin{itemize}[]
    \item We propose \obv, the first framework that integrates FL and targeted unlearning for LLMs, formulating them as a joint dual-objective optimization task to enable privacy-preserving training and compliance with GDPR’s \textit{right to be forgotten}.
    \item We consolidate diverse FL and unlearning benchmarks, training, and evaluation datasets into a user-friendly platform, facilitating standardized research for the LLM and FL communities.
    \item Our empirical evaluation reveals that \obv outperforms local training, with federated methods delivering an average model utility 27.43\% higher than the best local training. This achievement strikes a robust balance between forgetting efficacy and model utility, while cross-comparisons of algorithms provide valuable insights for advancing future LLM development.

\end{itemize}

%% file: Section/2-RelatedWork.tex
\section{Related Work}

\begin{figure*}[t]
    \centering
    \includegraphics[width=\linewidth]{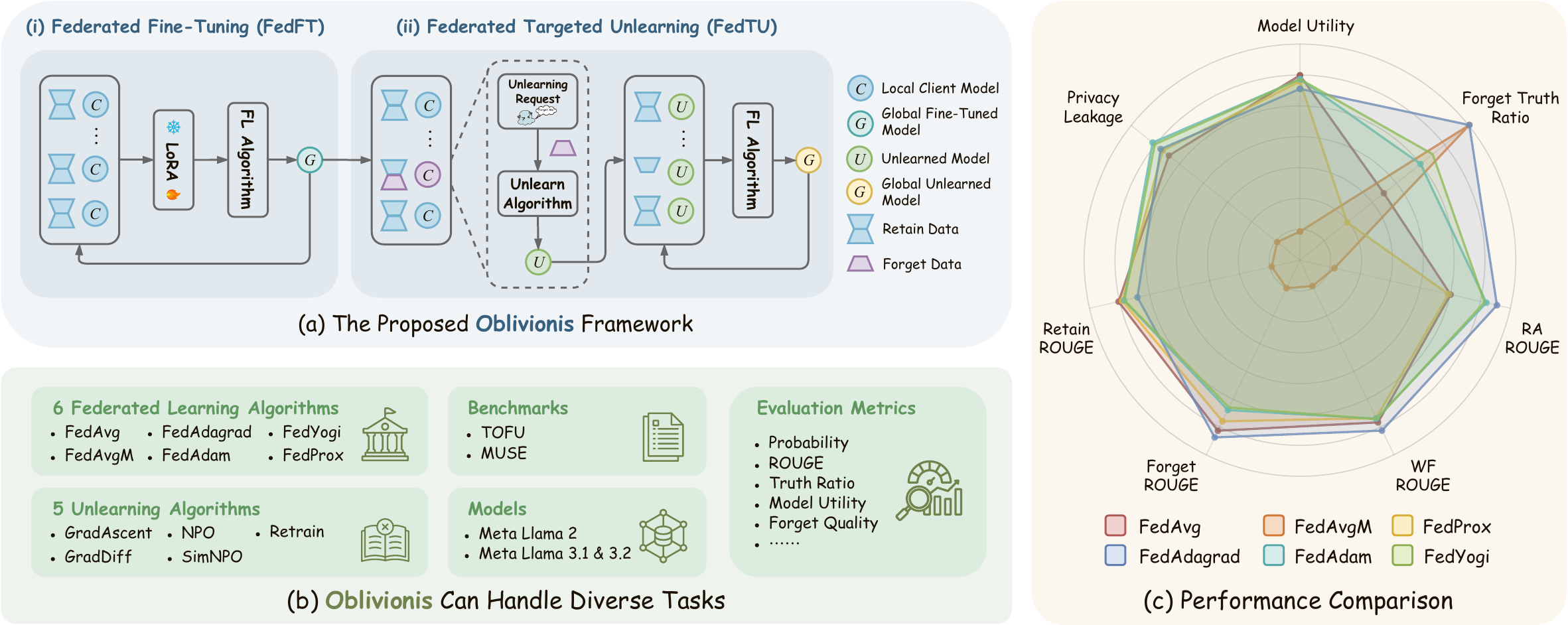}
    \caption{\textbf{(a)} Overview of the proposed \obv framework. {\textbf{(b)} \obv integrates 6 representative federated learning algorithms, 5 machine unlearning methods, 2 federated fine-tuning methods (full-parameter and LoRA-based), and a variety of models. \obv also supports 5 datasets and over 10 evaluation metrics. \textbf{(c)} Sample experimental results that showcase the divergent performance of \textbf{6 FL methods} using \textbf{SimNPO} unlearning algorithm on the \textbf{TOFU} dataset.}}
    \label{fig:main}
\end{figure*}
\subsection{Federated Fine-Tuning}
FL enables collaborative optimization of a shared model across distributed clients without exposing clients' private training data to preserve privacy. Recent advancements in FL have been expressed by FedLLM frameworks. Chen~\etal~(\citeyear{chen2023federated}) propose a framework emphasizing pre-training, fine-tuning, and prompt engineering for privacy-sensitive applications in FedLLM. Fan~\etal~(\citeyear{fan2023fate}) introduce FATE-LLM, an industrial-grade framework with parameter-efficient fine-tuning and privacy mechanisms for enterprise usage. Ye~\etal~(\citeyear{ye2024openfedllm}) propose OpenFedLLM, enabling federated instruction tuning and value alignment, outperforming local training in financial benchmarks. Wu~\etal~(\citeyear{wu2024fedbiot}) present FedBiOT, a resource-efficient fine-tuning approach using compressed models and adapters. Wu~\etal~(\citeyear{wu2024client}) further explore federated Reinforcement Learning from Human Feedback (RLHF). They propose FedBis and FedBiscuit strategies to enhance FedLLM alignment while handling client preference heterogeneity~\cite{wu2024client}. These works significantly advance FedLLM training, enhancing efficiency and privacy for distributed learning. However, existing FedLLM frameworks often lack robust unlearning mechanisms, failing to address GDPR’s regulation or effectively remove low-quality or outdated data contributions.

\subsection{LLM Unlearning}
LLMs have achieved remarkable success across diverse domains, yet their dependence on enormous datasets raises significant privacy and ethical concerns, such as compliance with GDPR’s \textit{right to be forgotten} and the removal of low-quality knowledge or biased content. In response, machine unlearning has emerged as a critical mechanism to address these issues by selectively removing specific knowledge from trained models without compromising overall model performance. It strategically modifies the trained model to erase required information without retraining from scratch.

Dorna~\etal~(\citeyear{openunlearning2025}) introduce a unified framework to standardize and accelerate the evaluation of unlearning algorithms for large language models, ensuring reproducibility and transparency through consistent metrics and datasets. Yao~\etal~(\citeyear{yao2024large}) provide a comprehensive overview of LLM unlearning, highlighting challenges like catastrophic forgetting and the difficulty of unlearning deeply integrated knowledge. Liu~\etal~(\citeyear{liu2025rethinking}) reconsider LLM unlearning objectives from a gradient perspective, advocating algorithms that minimize the influence of target data on model gradients. To enhance efficiency, Jia~\etal~(\citeyear{jia2024soul}) introduce SOUL, leveraging second-order optimization to achieve faster convergence in unlearning tasks. Similarly, Ji~\etal~(\citeyear{ji2024reversing}) develop a framework based on logit differences, reversing forget-retain objectives to efficiently remove specific knowledge. More targeted approaches, such as UIPE by Wang~\etal~(\citeyear{wang2025uipe}), focus on disentangling knowledge related to forgetting targets, while Fan~\etal~(\citeyear{fan2024simplicity}) demonstrate that simpler negative preference optimization can also outperform. These works collectively highlight the diversity of approaches in LLM unlearning, ranging from gradient-based algorithms and second-order optimization to targeted knowledge removal and simplified objectives. Despite these advancements, existing frameworks rarely address the joint optimization of federated fine-tuning and unlearning, leaving a gap in achieving both forgetting and model utility, which \obv aims to fill.



%% file: Section/3-Overview.tex
\section{Overview of Framework}
This section formalizes the \obv framework. In \obv, multiple clients train a shared model collaboratively while enabling targeted removal of specific data contributions via unlearning requests, as shown in Figure~\ref{fig:main}. The framework treats FL and unlearning as a dual optimization problem, with FL denoted by the operator \( \mathcal{F}\) and unlearning by \( \mathcal{U}\), allowing flexibility for various methods.

\subsection{Federated Learning Setup}
Consider \( K \) clients in an FL framework, indexed by \( k \in \{1, 2, \dots, K\} \). Each client \( \mathcal{C}_k \) holds a private dataset: \( \mathcal{D}_k = \left\{\left(x_i, y_i\right)\right\}_{i=1}^{N_k} \), where \( x_i \) and \( y_i \) are sequences of tokens (input/prompt and output/response, respectively), and \( N_k = |\mathcal{D}_k| \) is the number of samples for client $k$. The token sequences are used to fine-tune an LLM parameterized by \( \theta \in \mathbb{R}^d \), where \( d \) is the dimensionality of the model parameters. Let \( y_{i,j} \) denote the \( j \)-th token in \( y_i \) given the concatenated sequence of input \( x_i \) and previous tokens \( y_{i,<j} = (y_{i,1}, \dots, y_{i,j-1}) \). The probability of generating \( y_{i,j} \) is \( p(y_{i,j} \mid x_i \oplus y_{i,<j}; \theta) \), where \( \oplus \) is the sequence concatenation operator.

To address the high communication overhead of full fine-tuning in FL, where transmitting the entire set of model parameters across clients is computationally expensive, we adopt Low-Rank Adaptation (LoRA)~\cite{hu2022lora} for parameter-efficient fine-tuning. LoRA achieves performance comparable to full fine-tuning while significantly reducing the communication and computational costs by updating only a small subset of parameters. Specifically, for each client \( \mathcal{C}_k \) at communication round \( t \in \{1, 2, \dots, T\} \), LoRA updates a subset of the model parameters for a given weight matrix \( \mathbf{W} \in \mathbb{R}^{m \times n} \) in the large language model  through a low-rank decomposition:
\begin{equation}
    \mathbf{W}_k^{t} = \mathbf{W} + \Delta \mathbf{W}_k^{t}, \quad \Delta \mathbf{W}_k^{t} = \mathbf{A}_k^{t} \mathbf{B}_k^{t}
\end{equation}

\noindent where \( \mathbf{A}_k^{t} \in \mathbb{R}^{m \times r} \), \( \mathbf{B}_k^{t} \in \mathbb{R}^{r \times n} \), and \( r \ll \min(m, n) \) is the rank of the adaptation. The global model parameters \( \theta_t \) include the fixed base weights \( W \), while each client \( \mathcal{C}_k \) optimizes the LoRA parameters \( \phi_k^{\left(t\right)} = \left\{\mathbf{A}_k^{t}, \mathbf{B}_k^{t}\right\} \) during local training. The full set of model parameters is denoted as \( \theta = \theta_{\text{base}} + \phi \), where \( \theta_{\text{base}} \) is the set of frozen pre-trained parameters, and \( \phi \) represents the LoRA parameters. Since \( r \) is small, \( |\phi| \ll |\theta| \), substantially reduces the parameter optimization burden.

\subsection{Federated Fine-Tuning (FedFT)}
Federated fine-tuning collaboratively optimizes the global model \( \theta_t \) across all clients over \( T \) communication rounds. Each client $k$ first conducts local training on its local model. The base model parameters \( \theta_{\text{base}} \) remain fixed. At round \( t \), client \( \mathcal{C}_k \) receives global LoRA parameters \( \phi^{t-1} \), initializes the local LoRA parameters \( \phi_k^{(t,0)} = \phi^{t-1} \) and performs \( R \) iterations of local optimization on \( \mathcal{D}_k \) using stochastic gradient descent (SGD) on the LoRA parameters. For iteration \( r \in \{1, 2, \dots, R\} \): 
\begin{equation}
\phi_k^{(t,r)} = \phi_k^{(t,r-1)} - \eta \nabla_{\phi} \mathcal{L}_k\left(\phi_k^{\left(t,r-1\right)}; \mathcal{B}_k\right)
\end{equation}

\noindent where \( \eta \) is the learning rate, and \( \mathcal{B}_k \subseteq \mathcal{D}_k \) is a mini-batch. The mini-batch loss is:

\begin{align}
\mathcal{L}_k\left(\phi_k^{\left(t,r-1\right)}; \mathcal{B}_k\right) = \frac{1}{|\mathcal{B}_k|} \sum_{(x_i, y_i) \in \mathcal{B}_k} \notag \\
\quad - \sum_{j=1}^{n_i} \log p \left( y_{i,j} \mid x_i \oplus y_{i,<j}; \theta_{\text{base}} + \phi_k^{\left(t,r-1\right)} \right)
\end{align}

\noindent where \( n_i = |y_i| \) is the length of the output sequence, and the probability is computed using the model with parameters \( \theta_{\text{base}} + \phi_k^{(t,r-1)} \). 

\noindent\textbf{Federated Learning Process:} The FL operator \( \mathcal{F} \) aggregates local updates to produce the global parameters:
\begin{equation}
\phi^t = \mathcal{F}(\{\phi_k^{(t,R)}\}_{k=1}^K, \phi^{t-1}, \{\mathcal{D}_k\}_{k=1}^K)
\end{equation}
where \( \mathcal{F} \) can represent methods like weighted averaging (e.g., \( \phi^t = \phi^{t-1} + \sum_{k=1}^K w_k (\phi_k^{(t,R)} - \phi^{t-1}) \), with \( w_k = \frac{N_k}{\sum_{j=1}^K N_j} \)) or other schemes. The FedFT objective is:
\begin{equation}
\mathcal{L}_{\text{FedFT}}(\phi^t) = \sum_{k=1}^K w_k \mathcal{L}_k(\phi^t; \mathcal{D}_k)
\end{equation}

\subsection{Federated Targeted Unlearning (FedTU)}
A client \( \mathcal{C}_u \in \{1, 2, \dots, K\} \) requests unlearning of a subset \( \mathcal{D}_u^{\text{forget}} = \{(x_i, y_i)\}_{i \in \mathcal{I}_u} \subseteq \mathcal{D}_u \), where \( \mathcal{I}_u \) is the index set of the data points to be unlearned. The goal is to derive global LoRA parameters \( \phi^t_{\text{unlearn}} \) that approximate a model trained without \( \mathcal{D}_u^{\text{forget}} \), while preserving performance on the remaining data \( \bigcup_{k=1}^K \mathcal{D}_k \setminus \mathcal{D}_u^{\text{forget}} \). The unlearning operator \( \mathcal{U} \) produces:
\begin{equation}
\phi^t_{\text{unlearn}} = \mathcal{U}(\phi^t, \mathcal{I}_u, \mathcal{D}_u^{\text{forget}})
\end{equation}
where \( \mathcal{U} \) represents a general unlearning method (e.g., gradient ascent, influence functions). The server updates the global parameters: \( \phi^{t+1} = \phi^t_{\text{unlearn}} \), and broadcasts \( \phi^{t+1} \) to all clients. Clients then resume local fine-tuning using Equation (2) to compute \(\phi_k^{(t+1,r)}\).
\begin{equation}
\phi^{t+2} = \mathcal{F}(\{\phi_k^{(t+1,R)}\}_{k=1}^K, \phi^{t+1}, \{\mathcal{D}_k \setminus \mathcal{D}_u^{\text{forget}}\}_{k=1}^K)
\end{equation}

The FedTU objective minimizes the influence of \( \mathcal{D}_u^{\text{forget}} \):
\begin{equation}
\mathcal{L}_{\text{FedTU}}(\phi^t_{\text{unlearn}}) = \mathcal{L}_{\text{FedFT}}(\phi^t_{\text{unlearn}}; \bigcup_{k=1}^K \mathcal{D}_k \setminus \mathcal{D}_u^{\text{forget}})
\end{equation}

Finally, the Unified Framework alternates between FedFT and FedTU, solving the dual optimization objective problem:
\begin{equation}
\min_{\phi^t_{\text{unlearn}}}\min_{\phi^t} \left( \mathcal{L}_{\text{FedFT}}(\phi^t)+\mathbb{I}_{\text{unlearn}}(t)\cdot\mathcal{L}_{\text{FedTU}}(\phi^t_{\text{unlearn}}) \right)
\end{equation}

This is achieved by iteratively applying \( \mathcal{F} \) for FedFT and \( \mathcal{U} \) for FedTU. At each communication round $t$, the server checks for unlearning requests from a client $\mathcal{C}_u$ specifying $\mathcal{D}_{u}^{\text{forget}}$. If present, \( \mathcal{U} \) is activated $\mathbb{I}_{\text{unlearn}}(t)=1$; otherwise, only \( \mathcal{F} \) is applied $\mathbb{I}_{\text{unlearn}}(t)=0$. Our framework supports various implementations of \( \mathcal{F} \)  and \( \mathcal{U} \), ensuring flexibility.

%% file: Section/4-Experiment.tex
\section{Experiments}

\subsection{Experimental Setups}
To explore the performance of different algorithms in the \obv framework, we conduct comprehensive experiments using a carefully designed experimental setup.

\subsubsection{Models and Benchmark Datasets.} 
We consider four base models in our experiments: \textit{Llama-2-7b-hf}~\cite{touvron2023llama}, \textit{Llama-3.1-8B-Instruct}, \textit{Llama-3.2-1B-Instruct}, and \textit{Llama-3.2-3B-Instruct}~\cite{grattafiori2024llama}. We fine-tune and evaluate these models on two benchmark datasets: \textbf{TOFU} and \textbf{MUSE}, selected based on prior works~\cite{wang2024llm, yuan2024closer, openunlearning2025}. The \textbf{TOFU} dataset is divided into four subsets: \textit{Forget Set (Forget)}, \textit{Retain Set (Retain)}, \textit{Real Authors (RA)}, and \textit{World Facts (WF)}. The \textbf{MUSE} dataset comprises two corpora, \textit{News} and \textit{Books}, to simulate real-world large-scale unlearning requests and evaluate forgetting efficacy and model utility preservation in machine unlearning algorithms. 

\subsubsection{Baselines.} We employ six well-established federated optimization algorithms and five unlearning algorithms as baselines, detailed as follows:

\begin{itemize}
    \item \textbf{FL Algorithms:} We categorize the considered FL algorithms into two groups: \textbf{Adaptive Optimization FL (AOFL)}, including \textit{FedAdagrad}, \textit{FedAdam}, and \textit{FedYogi}~\cite{reddi2020adaptive}, which enhance aggregation with momentum or adaptive learning rates; and \textbf{Weighted Averaging-Based FL (WAFL)}, comprising \textit{FedAvg}~\cite{mcmahan2017communication}, \textit{FedAvgM}~\cite{hsu2019measuring}, and \textit{FedProx}~\cite{li2020federated}, which focus on parameter averaging or regularization. By focusing on these foundational and widely applicable algorithms, \obv ensures scalability and extensibility for diverse FL scenarios.
    \item \textbf{Unlearning Algorithms:} Integrated unlearning algorithms are classified into two types: \textbf{Gradient-Based Optimization Unlearning (GOUL)}, including \textit{GradAscent}, \textit{GradDiff}~\cite{maini2024tofu}, and \textit{RMU}~\cite{li2024wmdp}; and \textbf{Preference Optimization Unlearning (POUL)}, including \textit{NPO}~\cite{zhang2024negative} and \textit{SimNPO}~\cite{fan2024simplicity}. \textbf{GOUL} directly manipulates gradients or representations to eliminate the influence of data targeted for forgetting, employing simpler, targeted adjustments. 
\end{itemize}

\vspace{-0.1cm}
\begin{table}[htp]
    \centering
    \resizebox{\linewidth}{!}{
    \begin{tabular}{ccccc}
        \toprule
        \textbf{Model} & \textbf{Size} & \textbf{$N_\text{Base}$} & \textbf{$N_\text{Trainable}$} & \textbf{Ratio (\%)} \\
        \midrule
        Llama-2         & 7B  & 6818.37 M  & 79.95 M  & 1.17 \\
        Llama-3.1       & 8B  & 8114.15 M & 83.89 M  & 1.03 \\
        \multirow{2}{*}{Llama-3.2} & 1B  & 1258.36 M & 22.54 M  & 1.79 \\
                                  & 3B  & 3261.38 M & 48.63 M & 1.49 \\
        \bottomrule
    \end{tabular}
    }
    \caption{Illustration of model parameter distribution.}
    \vspace{-0.2cm}
    \label{tab:lora_efficiency}
\end{table}

\begin{table*}[!htp]
  \centering
  \tiny
  \renewcommand{\arraystretch}{0.6}
  \resizebox{\textwidth}{!}{%

  \begin{tabular}{ccccccccccccc}
  \toprule
    \multirow{3}[3]{*}{\textbf{Algorithms}} & \multicolumn{6}{c}{\textbf{Weighted Averaging-Based FL}} & \multicolumn{6}{c}{\textbf{Adaptive Optimization FL}} \\
     \cmidrule(lr){2-7} \cmidrule(lr){8-13}
     & \multicolumn{2}{c}{\textbf{FedAvg}} & \multicolumn{2}{c}{\textbf{FedAvgM}} & \multicolumn{2}{c}{\textbf{FedProx}} & \multicolumn{2}{c}{\textbf{FedAdagrad}} & \multicolumn{2}{c}{\textbf{FedAdam}} & \multicolumn{2}{c}{\textbf{FedYogi}} \\
     \cmidrule(lr){2-3} \cmidrule(lr){4-5} \cmidrule(lr){6-7} \cmidrule(lr){8-9} \cmidrule(lr){10-11} \cmidrule(lr){12-13}
     & MU$\uparrow$ & FTR$\uparrow$ & MU$\uparrow$ & FTR$\uparrow$ & MU$\uparrow$ & FTR$\uparrow$ & MU$\uparrow$ & FTR$\uparrow$ & MU$\uparrow$ & FTR$\uparrow$ & MU$\uparrow$ & FTR$\uparrow$\\
    \toprule
    \rowcolor{cyan!5}
    \multicolumn{13}{c}{\textbf{Meta Llama-3.2-1B-Instruct with LoRA}} \\
    \midrule
    \rowcolor{gray!10} \textbf{Finetune} & 0.50 & 0.49 & 0.48 & 0.45 & 0.50 & 0.49 & 0.45 & 0.62 & 0.45 & 0.60 & 0.45 & 0.59 \\
    \textbf{GradAscent} & \cellcolor{cyan!20}{\textbf{0.46}} & 0.61 & 0 & 0.050 & 0.43 & 0.64 & 0.40 &  \underline{0.72} & 0.44 & 0.65 & \cellcolor{cyan!20}{\textbf{0.46}} & 0.66 \\
    \textbf{GradDiff} & \cellcolor{cyan!20}{\textbf{0.46}} & 0.63 & 6.5e-5 &  \underline{0.70} & 0.44 & 0.60 & 0.42 &  \underline{0.70} & 0.44 & 0.66 & 0.44 & 0.67 \\
    \textbf{NPO} & \cellcolor{cyan!20}{\textbf{0.46}} & 0.62 & 2.9e-5 & 0.71 & 0.44 & 0.63 & 0.41 & \underline{0.74} & 0.45 & 0.68 & 0.45 & 0.68 \\
    \textbf{SimNPO} & \cellcolor{cyan!20}{\textbf{0.46}} & 0.65 & 0.00018 & 0.69 & 0.43 & 0.66 & 0.42 & \underline{0.74} & \cellcolor{cyan!20}{\textbf{0.46}} & 0.69 & \cellcolor{cyan!20}{\textbf{0.46}} & 0.70 \\
    \textbf{Retrain} & \cellcolor{cyan!20}{\textbf{0.51}} & 0.65 & 0.47 & 0.62 & \cellcolor{cyan!20}{\textbf{0.51}} & 0.64 & 0.46 &  \underline{0.67} & 0.46 & 0.66 & 0.46 & 0.66 \\
    \midrule
    \rowcolor{cyan!5}
    \multicolumn{13}{c}{\textbf{Meta Llama-3.2-3B-Instruct with LoRA}} \\
    \midrule
    \rowcolor{gray!10} \textbf{Finetune} & 0.59 & 0.49 & 0.56 & 0.48 & 0.58 & 0.51 & 0.53 & 0.61 & 0.50 & 0.57 & 0.50 & 0.57 \\
    \textbf{GradAscent} & \cellcolor{cyan!20}{\textbf{0.52}} & 0.59 & 0.00015 & \underline{0.79} & 0.48 & 0.62 & 0.45 & 0.73 & \cellcolor{cyan!20}{\textbf{0.52}} & 0.66 & 0.51 & 0.66 \\
    \textbf{GradDiff} & \cellcolor{cyan!20}{\textbf{0.52}} & 0.59 & 0.00062 &  \underline{0.77} & 0.49 & 0.59 & 0.47 & 0.71 & 0.51 & 0.61 & 0.51 &0.61\\
    \textbf{NPO} & \cellcolor{cyan!20}{\textbf{0.50}} & 0.62 & 0.00032 & \underline{0.79} & 0.47 & 0.60 & 0.45 & 0.73 & \cellcolor{cyan!20}{\textbf{0.50}} & 0.63 & \cellcolor{cyan!20}{\textbf{0.50}} &0.63\\
    \textbf{SimNPO} & \cellcolor{cyan!20}{\textbf{0.51}} & 0.61 & 0.0013 &  \underline{0.77} & 0.48 & 0.62 & 0.47 & 0.73 & 0.50 & 0.63 & \cellcolor{cyan!20}{\textbf{0.51}} & 0.65\\
    \textbf{Retrain} & \cellcolor{cyan!20}{\textbf{0.59}} & 0.64 & 0.56 & 0.64 & 0.57 & 0.65 & 0.53 &  \underline{0.66} & 0.50 & 0.63 & 0.50 & 0.63\\
    \bottomrule
  \end{tabular}}
  \caption{Performance comparison of federated learning and unlearning algorithms on the \textbf{TOFU} dataset using \textbf{Llama-3.2-1B and 3B} models, evaluated on metrics MU (Model Utility) and FTR (Forget Truth Ratio) with \textbf{Split99} strategies. Scores in \cellcolor{cyan!20}{\textbf{Bold}} indicate the optimal MU in different FL methods, while scores \underline{underlined} indicate the optimal FTR in different FL methods.}
  \label{tab:federated_learning_classic_layout}
\end{table*}

\begin{figure*}[!ht]
    \centering
    \includegraphics[width=1.0\linewidth]{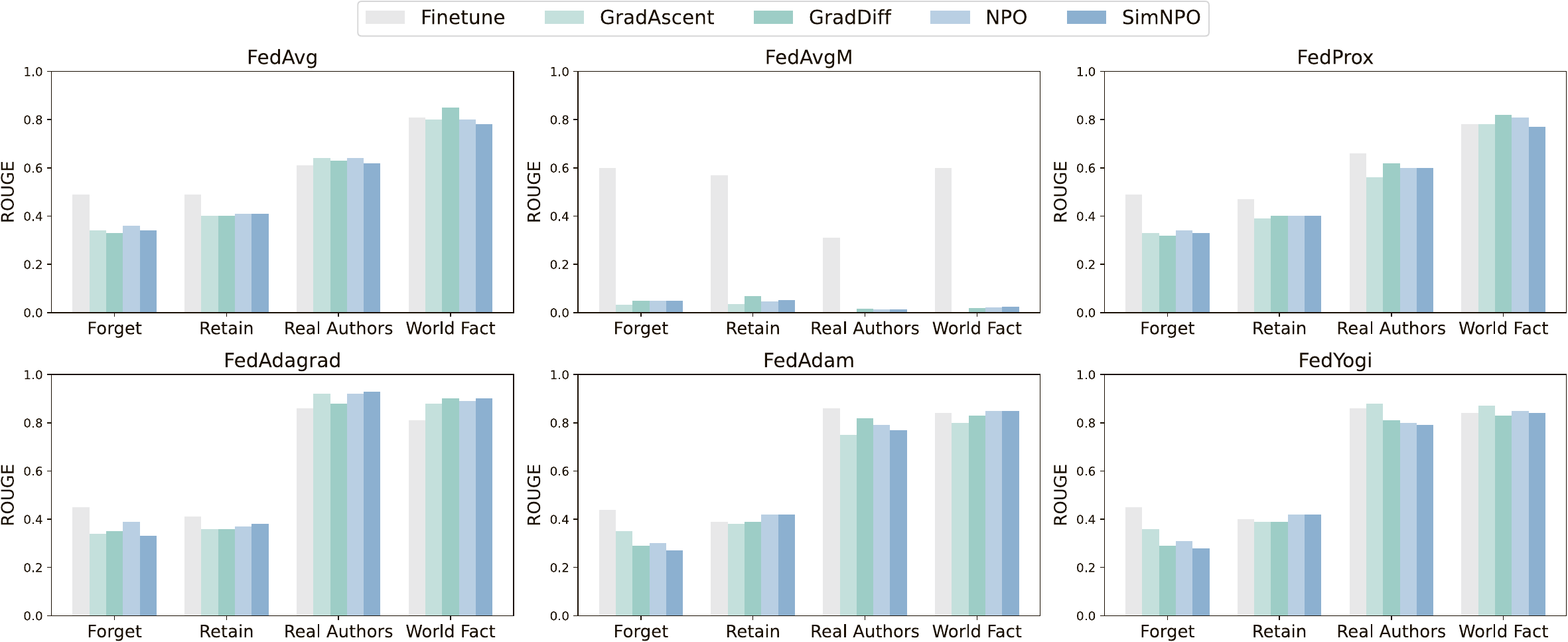}
    \caption{Comparative analysis of \textbf{ROUGE} scores across federated learning and unlearning methods using \textbf{Llama-3.2-1B} model with \textbf{Split99} strategies. For the Forget set, lower scores indicate better performance ($\downarrow$), whereas for the remaining sets, higher scores are preferable ($\uparrow$).}
    \label{fig:rouge_score}
\end{figure*}

\subsubsection{Training Setup.} We conduct experiments using 30 clients with a $10\%$ participation rate for \obv. In each round, a randomly selected client requests targeted sample-level unlearning. The training process consists of 5 local epochs and 10 global rounds, with a one-epoch warmup period included. The base models are fine-tuned using LoRA with a rank of 32, an alpha of 64, and a dropout rate of 0.05. We train the model with a learning rate of $8 \times 10^{-5}$ and a weight decay of 0.01. The entire experiment is tested on a cloud server with one NVIDIA A100 (80 GB) GPU. 

Meanwhile, Table~\ref{tab:lora_efficiency} summarizes the number of trainable parameters under the LoRA paradigm. In all cases, no more than 1.79\% of base models’ parameters are updated, while the rest remain frozen, highlighting the lightweight nature of our approach. For more experimental settings, including specific methods of federated learning and unlearning, datasets, and models, please refer to the contents in Appendix~\ref {Details} and~\ref {Experimental}.

\subsection{Experimental Results} 

\begin{figure*}[!ht]
    \centering
    \includegraphics[width=1.0\linewidth]{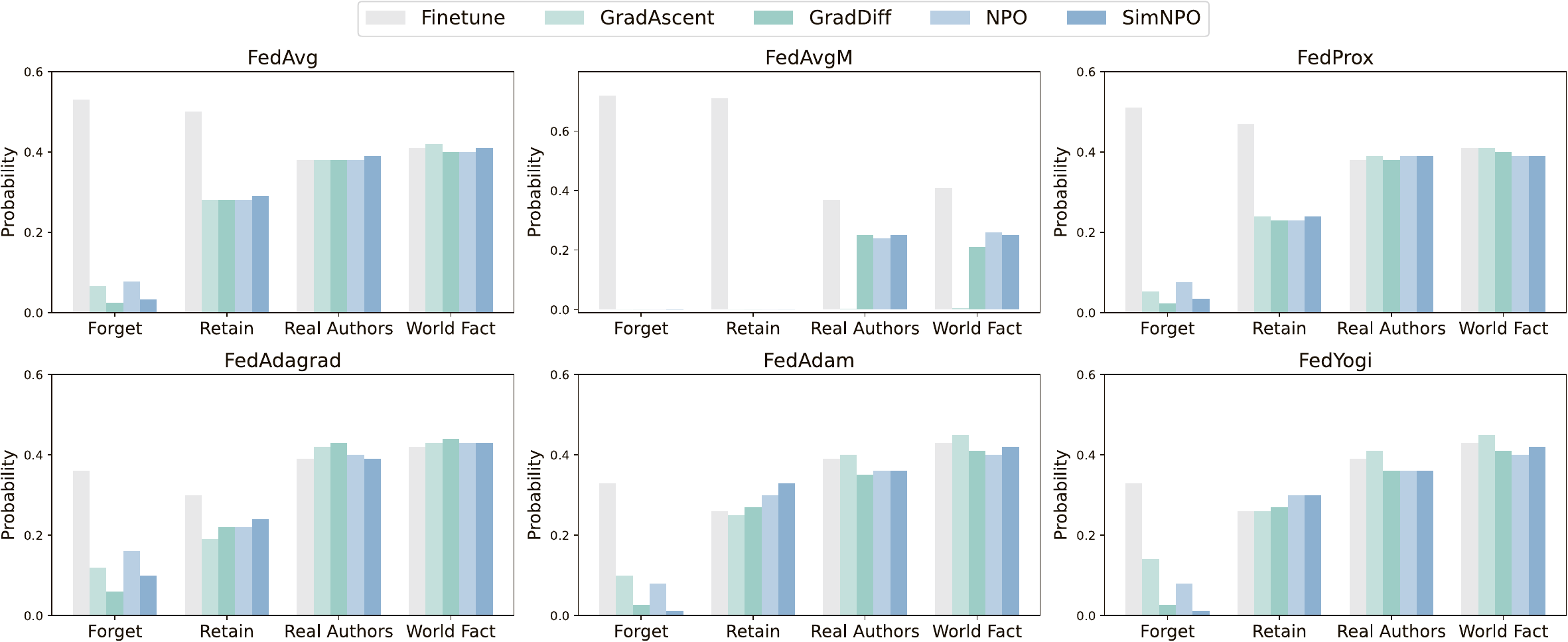}
    \caption{Comparative analysis of \textbf{Probability} scores across federated learning and unlearning methods using \textbf{Llama-3.2-1B} model with \textbf{Split99} strategies. For the Forget set, lower scores indicate better performance ($\downarrow$), whereas for the remaining sets, higher scores are preferable ($\uparrow$).}
    \label{fig:prob}
\end{figure*}

\subsubsection{Structured QA Task.}



As presented in Table~\ref{tab:federated_learning_classic_layout}, we choose Model Utility (MU) and Forget Truth Ratio (FTR) to evaluate. AOFL algorithms, particularly FedAdagrad, consistently outperform WAFL methods in forgetting efficacy. For the 1B model, FedAdagrad, when paired with SimNPO or NPO, achieves an FTR of $0.74$, surpassing FedAvg's $0.65$ and FedProx's $0.66$. Similarly, for the 3B model, FedAdagrad attains an FTR of $0.73$, compared to $0.64$ for FedAvg and $0.65$ for FedProx. These findings indicate that AOFL methods effectively utilize adaptive optimization to prioritize the Forget Set objectives, thereby maintaining unlearning performance. However, this enhancement results in a reduction in MU, with FedAdagrad yielding MU values ranging from $0.40$ to $0.47$, whereas FedAvg maintains more stable MU values between $0.46$ and $0.59$ across both models. Among unlearning strategies, SimNPO and NPO demonstrate superior forgetting efficacy, achieving FTR values between $0.69$ and $0.74$ with AOFL methods while maintaining competitive MU values from $0.42$ to $0.51$. In contrast, the Retrain strategy achieves the highest MU value of up to $0.59$ but is computationally intensive, limiting its practical applicability. \textbf{Meanwhile, FedAvgM suffers from catastrophic forgetting in the Structured QA Task}, with MU values plummeting to between $0.00018$ and $0.0013$, despite achieving high FTR values of up to $0.79$. This instability likely arises from FedAvgM amplifying the adverse effects of unlearning updates on general model parameters, resulting in performance collapse.

To evaluate the impact of model scale, we test the larger 3B model, which shows higher MU and FTR values, indicating a better balance between utility and forgetting. For instance, FedAvg with Retrain achieves a MU of $0.59$ and an FTR of $0.64$ for the 3B model, compared to $0.51$ and $0.65$ for the 1B model. WAFL methods like FedAvg and FedProx yield stable MU values of $0.47$ to $0.59$ but lag in FTR compared to AOFL methods. This highlights a trade-off: AOFL methods prioritize forgetting but reduce utility, while WAFL methods ensure stability. All unlearning strategies except Finetune outperform the Finetune baseline’s FTR of $0.45$ to $0.62$ for the 1B model and $0.48$ to $0.61$ for the 3B model, achieving values of $0.59$ to $0.79$, confirming \obv’s robust unlearning capability.


\begin{table*}[ht]
\centering
\resizebox{\textwidth}{!}{
\begin{tabular}{ccccccccccccccccccc}
\toprule
\multirow{3}[3]{*}{\textbf{Algorithms}} & \multicolumn{9}{c}{\textbf{Weighted Averaging-Based FL}} & \multicolumn{9}{c}{\textbf{Adaptive Optimization FL}} \\
\cmidrule(lr){2-10} \cmidrule(lr){11-19}
& \multicolumn{3}{c}{\textbf{FedAvg}} & \multicolumn{3}{c}{\textbf{FedAvgM}} & \multicolumn{3}{c}{\textbf{FedProx}} & \multicolumn{3}{c}{\textbf{FedAdagrad}} & \multicolumn{3}{c}{\textbf{FedAdam}} & \multicolumn{3}{c}{\textbf{FedYogi}} \\ 
\cmidrule(lr){2-4} \cmidrule(lr){5-7} \cmidrule(lr){8-10} \cmidrule(lr){11-13} \cmidrule(lr){14-16} \cmidrule(lr){17-19}

& \multicolumn{1}{c}{NVM} & \multicolumn{1}{c}{NKM} & \multicolumn{1}{c}{UP} & \multicolumn{1}{c}{NVM} & \multicolumn{1}{c}{NKM} & \multicolumn{1}{c}{UP} & \multicolumn{1}{c}{NVM} & \multicolumn{1}{c}{NKM} & \multicolumn{1}{c}{UP} & \multicolumn{1}{c}{NVM} & \multicolumn{1}{c}{NKM} & \multicolumn{1}{c}{UP} & \multicolumn{1}{c}{NVM} & \multicolumn{1}{c}{NKM} & \multicolumn{1}{c}{UP} & \multicolumn{1}{c}{NVM} & \multicolumn{1}{c}{NKM} & \multicolumn{1}{c}{UP} \\ 
\midrule

\rowcolor{gray!10} \textbf{Finetune} & 0.77 & 0.57 & 0.43 & 0.34 & 0.38 & 0.31 & 0.60 & 0.60 & 0.52 & 0.61 & 0.65 & 0.53 & 0.67 & 0.63 & 0.50 & 0.67 & 0.62 & 0.50 \\
\textbf{GradAscent} & 0.41 & 0.49 & 0.35 & \underline{0.0059} & 0.030 & 0.019 & 0.56 & 0.56 & \cellcolor{cyan!20}{\textbf{0.49}} & 0.033 & \underline{0} & 0 & 0.46 & 0.51 & 0.40 & 0.44 & 0.50 & 0.41 \\
\textbf{GradDiff} & 0.39 & 0.43 & 0.34 & 0.25 &  \underline{0.24} & 0.24 & 0.52 & 0.55 & \cellcolor{cyan!20}{\textbf{0.52}} &  \underline{0.17} & 0.53 & 0.43 & 0.46 & 0.49 & 0.39 & 0.43 & 0.53 & 0.38 \\
\textbf{NPO} & 0.36 & 0.45 & 0.35 &  \underline{0.33} &  \underline{0.38} & 0.34 & 0.42 & 0.56 & \cellcolor{cyan!20}{\textbf{0.43}} & 0.36 & 0.50 & 0.36 & 0.39 & 0.47 & 0.35 & 0.43 & 0.44 & 0.39 \\
\textbf{SimNPO} & 0.32 &  \underline{0.39} & 0.33 & 0.30 & 0.41 & 0.29 & 0.27 & 0.51 & \cellcolor{cyan!20}{\textbf{0.42}} &  \underline{0.18} & 0.49 & 0.36 & 0.31 & 0.45 & 0.36 & 0.33 & 0.47 & 0.38 \\
\textbf{Retrain} & 0.21 & 0.32 & 0.46 &  \underline{0.18} &  \underline{0.22} & 0.30 & 0.21 & 0.36 & \cellcolor{cyan!20}{\textbf{0.52}} & 0.21 & 0.33 & \cellcolor{cyan!20}{\textbf{0.52}} & 0.21 & 0.32 & 0.50 & 0.21 & 0.34 & 0.50 \\
\bottomrule
\end{tabular}
}
\caption{Performance comparison of federated learning algorithms on the \textbf{MUSE News} set using \textbf{Llama-2-7B model}, evaluated on metrics NVM (No Verbatim Mem$\downarrow$), NKM (No Knowledge Mem$\downarrow$), and UP (Utility Preserved$\uparrow$). Scores in \cellcolor{cyan!20}{\textbf{Bold}} indicate the optimal UP in different FL methods, while \underline{underlined} indicate the optimal NVM and NKM in different FL methods.}
\label{tab:muse_fed_news}
\end{table*}

To validate the effectiveness of \obv in forgetting and retaining general knowledge, we evaluated it on all four sets from TOFU, using ROUGE and Probability metrics. These metrics analyze the model’s forgetting behavior from different perspectives: Forget ROUGE measures the textual similarity between generated and true answers in the Forget Set via ROUGE-L recall, indicating whether the model still produces targeted forgotten information; Forget Probability quantifies the conditional probability of correct answers, capturing subtle changes in output content and probability distribution. As shown in Figure~\ref{fig:rouge_score} and Figure~\ref{fig:prob}, on the Forget Set, from the initial fine-tuned model to each FU dual-optimization method, both Forget ROUGE and Forget Probability significantly decreased, indicating that the model’s generated answers deviated from the true answers, with a substantial reduction in probability preference for correct answers, proving the FU algorithm’s effectiveness in altering model output behavior and achieving information forgetting. Meanwhile, on the Retain Set, World Facts, and Real Authors sets, ROUGE and Probability results remained largely consistent with fine-tuning performance, demonstrating that the FU algorithm effectively retains model performance on non-forgotten data while forgetting the Forget Set. Overall, the evaluation confirms the FU algorithm’s effective capability for forgetting while maintaining the model’s overall performance stability. \textbf{Overall, FedAdagrad excels in forgetting efficacy but compromises model utility, whereas FedAvg and FedProx prioritize utility stability, sacrificing forgetting performance in the Structured QA Task.} For a comprehensive analysis involving various model scales and data split strategies, please refer to the results in Appendix~\ref{Supplementary Experiments}. 

\begin{figure}[!htp]
    \centering
    \includegraphics[width=\linewidth]{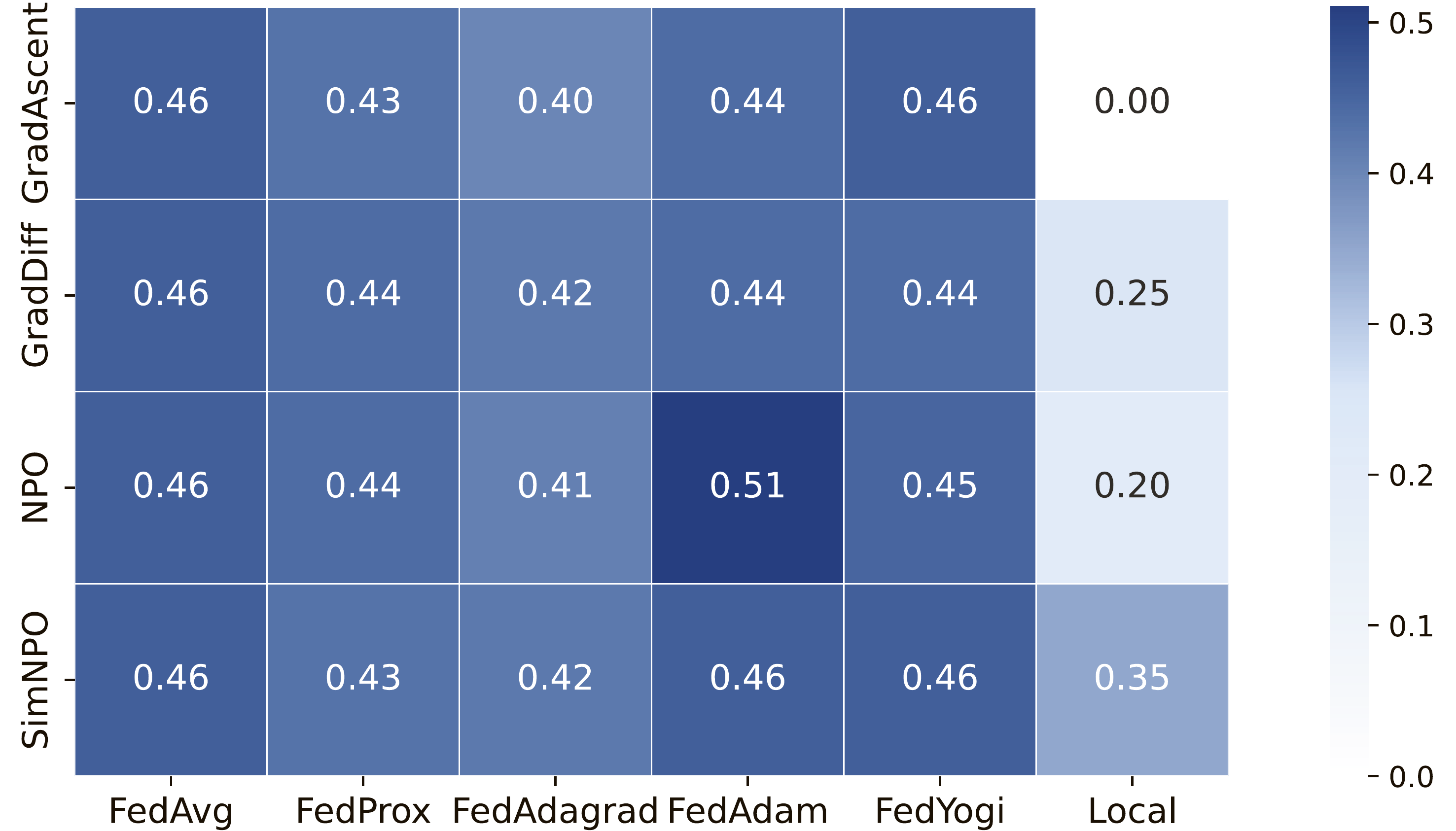}
    \caption{Comparison of \textbf{Model Utility(MU)} between local and federated learning across different unlearning methods.}
    \label{fig:heap_map}
\end{figure}

\subsubsection{Contextual QA Task.}

FedProx demonstrated a good balance across all objectives on the MUSE News set, as evidenced by Table~\ref{tab:muse_fed_news}. When combined with GradDiff, it achieves low NVM and NKM of 0.52 and 0.55, respectively, while maintaining high Utility Preserved (UP) at 0.52. These results indicate effective unlearning with robust model performance. FedAvg exhibits moderate performance. When combined with GradAscent, it yields an NVM of 0.41, NKM of 0.49, and UP of 0.35. These results indicate that it is less effective than FedProx in balancing forgetting and model utility. FedAvgM shows poor overall performance. For instance, when combined with GradAscent, it yields extremely low UP at 0.019, despite favorable NVM and NKM of 0.0059 and 0.03, respectively. Therefore, we consider it unsuitable for balanced optimization. Among the optimizer-enhanced methods, FedAdam and FedYogi delivered competitive performance. FedAdam achieves an NVM of 0.31, NKM of 0.45, and UP of 0.36 with SimNPO. FedYogi produces similar results with SimNPO, achieving an NVM of 0.33, NKM of 0.47, and UP of 0.38. FedAdagrad achieves less consistent results. When combined with GradDiff, it yields an NVM of 0.17, NKM of 0.53, and UP of 0.43. 

From a dual-objective optimization perspective, FedProx effectively minimizes NVM and NKM while maintaining high UP across all unlearning algorithms. FedAdam and FedYogi also achieve a well-balanced trade-off among the objectives, especially when combined with SimNPO. However, its effectiveness is slightly lower than that of FedProx. In contrast, FedAvg emphasizes model utility at the cost of unlearning performance, while FedAvgM prioritizes unlearning performance at the expense of model utility, making both approaches suboptimal. SimNPO and NPO demonstrate robust performance across FL methods, with SimNPO achieving the lowest NVM of 0.27 when paired with FedProx. In summary, \obv demonstrates strong effectiveness in balancing the dual-objective optimization of minimizing memorization, while maximizing utility across a majority of the scenarios considered. 
\textbf{Overall, FedProx demonstrates a better trade-off between model utility and unlearning performance in the contextual QA task.}

\subsubsection{Comparative Analysis of Local and Federated Learning.}
Empirical results illustrated in Figure~\ref{fig:heap_map} reveal that FU methods consistently achieve higher MU scores than local training across all unlearning strategies, demonstrating superior robustness in preserving model utility during unlearning. Local training exhibits a significant vulnerability to catastrophic forgetting, especially with GradAscent, where MU drops to near-zero levels. In contrast, FL methods mitigate the destabilizing effects of unlearning through collaborative parameter updates and maintain stable and competitive MU scores. Among the unlearning methods, NPO paired with FL algorithms yields the highest MU, indicating strong compatibility with the dual-objective optimization framework. In contrast, local training fails to balance unlearning and performance retention across all methods. \textbf{In summary, \obv significantly outperforms local training by maintaining robust model utility across unlearning methods, highlighting its efficacy for practical applications. }

%% file: Section/5-Conclusion.tex
\section{Conclusion}

In this work, we introduce \obv, a lightweight framework that seamlessly integrates federated learning and unlearning to enable distributed model training and compliance with regulations such as GDPR’s \textit{right to be forgotten}. By formulating FL and unlearning as a joint dual-objective optimization task, \obv achieves a robust balance between forgetting targeted data and preserving model utility, as demonstrated by superior performance on TOFU and MUSE benchmarks. Our comprehensive evaluation, including cross-comparisons of diverse FL and unlearning algorithms, evidences that models trained using \obv consistently outperform those trained using local training approaches. Notably, methods like FedAdagrad paired with SimNPO achieve high forgetting efficacy. By consolidating diverse benchmarks and datasets into a user-friendly code library, \obv further facilitates standardized research for the LLM and FL communities. Our framework is also open-sourced to facilitate reproducibility and foster further research in the development of LLM.

%% file: Section/Appendix.tex
\clearpage
\appendix

\section*{Appendix}
This appendix provides supplementary materials to facilitate additional insight into our \obv. It provides detailed descriptions of the benchmarks, evaluation metrics, models, and algorithms used in this framework. Additionally, we present implementation details, which cover the experimental setup, hyperparameters, and prompt templates. Further experimental results and a discussion of \obv limitations are also included.

\subsection*{Table of Contents}
\renewcommand{\labelitemi}{}
\renewcommand{\labelitemii}{}
\begin{itemize}
  \item \textbf{A} \quad \textbf{\obv Details} \hfill \pageref{Details}
  \begin{itemize}
      \item A.1 \quad Benchmarks \dotfill \pageref{Benchmarks}
      \item A.2 \quad Metrics \dotfill \pageref{Metrics}
      \item A.3 \quad Models \dotfill \pageref{Models}
      \item A.4 \quad Federated Methods \dotfill \pageref{Federated}
      \item A.5 \quad Unlearning Methods \dotfill \pageref{Unlearning}
  \end{itemize}
  \item \textbf{B} \quad \textbf{Experimental Details} \hfill \pageref{Experimental}
    \begin{itemize}
      \item B.1 \quad Testbed \dotfill \pageref{Testbed}
      \item B.2 \quad Hyperparameters \dotfill \pageref{Hyperparameters}
      \item B.3 \quad Prompt Template \dotfill \pageref{Prompt}
      \item B.4 \quad Overheads \dotfill \pageref{Overheads}
  \end{itemize}
  \item \textbf{C} \quad \textbf{Limitations} \hfill \pageref{Limitations}
  \item \textbf{D} \quad \textbf{Supplementary Experiments} \hfill \pageref{Supplementary Experiments}
\end{itemize}

\section{\obv Details}
\label{Details}
\subsection{Benchmarks}
\label{Benchmarks}

\obv includes multiple unlearning benchmarks, each designed to target specific aspects of forgetting in LLMs. Together, these benchmarks form a comprehensive testbed for evaluating unlearning methods under diverse scenarios.

\subsubsection{TOFU.}
TOFU (Task of Fictitious Unlearning), proposed by Maini~\etal~(\citeyear{maini2024tofu}), is a question-answer (QA)-format benchmark specifically designed to evaluate the unlearning capabilities of LLMs. It consists of QA pairs derived from autobiographies of 200 fictitious authors, generated entirely by GPT-4 to ensure the content does not exist in the pretraining corpora of existing LLMs. Each author profile includes 20 question-answer pairs, covering attributes such as name, birthplace, gender, birth year, genre, awards, and parents' occupations, with book titles seeded from the Goodreads Books dataset to enhance diversity. The benchmark is structured into four distinct sets: 1) \textit{Forget Set}: targeted for unlearning, comprising 1\%, 5\%, or 10\% of the data, corresponding to 2, 10, or 20 authors; 2) \textit{Retain Set}: data to be preserved, comprising 90\%, 95\%, or 99\% of the data; 3) \textit{Real Authors}: used to assess knowledge of real-world entities, and 4) \textit{World Facts}: used to evaluate general knowledge retention. This benchmark provides a controlled environment to study unlearning efficacy, providing a precise evaluation of a model's ability to forget specific information while maintaining performance on unrelated tasks.

\subsubsection{MUSE.}
MUSE, introduced by Shi~\etal~(\citeyear{shi2024muse}), is a comprehensive unlearning evaluation benchmark targeting the removal of articles from a fine-tuned LLM. It comprises two corpora: \textbf{News}, based on BBC news articles, and \textbf{Books}, based on the Harry Potter book series. The \textbf{News} corpus includes: \textit{Forget Set} (0.8M tokens) and \textit{Retain Set} (1.6M tokens) of disjoint news articles, while \textbf{Books} corpus designates the Harry Potter books (3.3M tokens) as the Forget Set and related Wikipedia articles as the Retain Set. Each corpus contains verbatim text and a knowledge set of question-answer pairs generated by GPT-4, with answers extracted verbatim from the text to assess memorization. A Holdout Set $\mathcal{D}_{\text{Holdout}}$ is included to evaluate privacy leakage, and a distinct Retain Set $  \mathcal{D}_{\text{Retain}}^{\text{(Reg)}}$ supports regularization during unlearning.


\subsection{Metrics}
\label{Metrics}

These metrics are broadly categorized into three types, as summarized below:

\subsubsection{Memorization Metrics.}

These metrics quantify the extent to which the model retains information from its training data.


\begin{itemize}
    \item \textbf{Probability:} We measure the likelihood of generating correct answers, reported as a probability in $[0, 1]$. We compute the conditional probability $P(a \mid q)$ according to the model for the Retain Set. Following standard practice~\cite{cho-etal-2014-properties}, we normalize for answer length by exponentiating the probability to the power of $1 / |a|$, as shown in Equation~\ref{eq: p_rf}:
    \begin{equation}
    \label{eq: p_rf}
    P_{\text{Retain}}(x) = P(a \mid q)^{1 / |a|}.
    \end{equation}
    
    For \textit{Real Authors} and \textit{World Facts}, we calculate the relative probability of the correct answer in a multiple-choice setting:
    \begin{equation}
        P_{\text{Real/World}}(x) = \frac{P(a_1 \mid q)}{\sum_{i=1}^n P(a_i \mid q)},
    \end{equation}
    where question $q$ is a multiple-choice question associated with choices $\{a_1, \dots, a_n\}$, with $a_1$ designated as the correct answer.

\item \textbf{Recall-Oriented Understudy for Gisting Evaluation (ROUGE):} We employ ROUGE scores to evaluate the similarity between model-generated answers and the ground truth. Specifically, we use ROUGE-L~\cite{lin2004rouge} recall, which measures similarity based on the longest common subsequence (LCS). It approximates the accuracy in the question-answering task, accounting for minor phrasing variations in output compared to the ground truth.

\begin{equation}
\text{ROUGE}_L(x) = \frac{\text{LCS}(a, \hat{a})}{|a|},
\end{equation}
where $\hat{a}$ is the generated answer, and $\text{LCS}(a, \hat{a})$ is the LCS length between $a$ and $\hat{a}$.

\item \textbf{Truth Ratio:} For a given question, we define the truth ratio $R_{\text{truth}}$ as an approximate comparison between the likelihood of the correct answer and that of incorrect answers. As the model is fine-tuned on a specific phrasing of the ground truth answer, its probability may be inflated relative to other formulations of the correct answer. Thus, we evaluate the probability of a paraphrased version of the answer instead of the original. Similarly, rather than comparing against a single incorrect answer, we compute the average probability of multiple incorrect answers formatted similarly to the paraphrased answer. 

Let $\tilde{a}$ denote the paraphrased correct answer, with $\tilde{x} = [q, \tilde{a}]$. $A_{\text{err}}$ is a set of incorrect answers $a_\text{err}$, modifying the answer to preserve the text’s general structure while introducing factual inaccuracies. The truth ratio $R_{\text{Truth}}$ is then defined as follows:

\begin{equation}
    R_{\text{Truth}}(x) = \frac{\frac{1}{|A_{\text{err}}|} \sum_{a_{\text{err}} \in A_{\text{err}}} P(a_{\text{err}} \mid q)^{1 / |a_{\text{err}}|}}{P(\tilde{a} \mid q)^{1 / |\tilde{a}|}}
\end{equation}

Additionally, as detailed in Equation~\ref{eq:R_adj}, we normalize and rescale the metrics to ensure that each value lies within the interval $[0, 1]$, where higher values correspond to better model performance. 

\begin{equation}
    \label{eq:R_adj}
    R_{\text{Adjusted}}(x) = \max(0, 1 - R_{\text{Truth}}(x)).
\end{equation}

\end{itemize}

\subsubsection{Privacy Metrics.} These metrics assess whether sensitive information from the forget set can still be inferred or extracted from the model. However, it is important to note that they often rely on idealized assumptions, such as access to perfectly i.i.d. holdout data or an oracle retain model, which may limit their applicability in real-world scenarios.

\begin{itemize}
    \item \textbf{Forget Quality: } 
    In our setting, we use $F_U(x)$ and $F_R(x)$ to denote the empirical cumulative distribution functions (CDFs) of the unlearned and retained models, constructed from $n$ and $m$ samples, respectively. The Kolmogorov–Smirnov (KS) test then computes the test statistic as follows:
\begin{equation}
    D_{n,m} = \sup_{x} \left| F_U(x) - F_R(x) \right|
\end{equation}

which measures the maximum deviation between two empirical distributions. This metric quantifies the distributional shift introduced by the unlearning process, enabling a non-parametric comparison between two models.

This metric quantifies the distributional shift introduced by the unlearning process, enabling a non-parametric comparison between the two models.
    
The null hypothesis (i.e., identical distributions), stating that the two sets of samples are drawn from the same distribution, is rejected at a given significance level $\alpha$ if the following inequality is satisfied:
\begin{equation}
    D_{n,m} > c(\alpha) \sqrt{\frac{n + m}{nm}}
\end{equation}
where $c(\alpha)$ denotes the critical value associated with the significance level $\alpha$, computed as
\begin{equation}
    c(\alpha) = \sqrt{-\frac{1}{2} \ln\left(\frac{\alpha}{2}\right)}
\end{equation}

The resulting $p$-value is defined as the smallest significance level $\alpha$ such that the null hypothesis can be rejected:

$$
Q_{\text{Forget}} = p = \min \left\{ \alpha \,\middle|\, D_{n,m} > c(\alpha) \sqrt{\frac{n + m}{nm}} \right\}
$$

Consequently, \textbf{Forget Quality} reflects the statistical confidence with which we can assert that the distributions of Truth Ratio values over the forget set from the unlearned and retained models are different.

\end{itemize}

\subsubsection{Utility Metrics.} The goal of unlearning is to effectively remove the influence of the targeted data while preserving the model's performance on non-forget data. Utility metrics evaluate whether the model maintains its capabilities on broader tasks beyond the retain set, thereby ensuring that unlearning does not compromise general performance on real-world distributions.

\begin{itemize}
    \item \textbf{Model Utility.} Model Utility (MU) captures the retained performance of a model after unlearning, both on the closely related retain set and on broader general knowledge. 
    
    TOFU evaluates Model Utility as the harmonic mean of nine metrics spanning three data levels: the retain set, real authors, and factual world knowledge to ensure balanced performance. At each level, it computes three metrics mentioned before: \textbf{Probability}, \textbf{ROUGE}, and \textbf{Truth Ratio}.
\begin{equation}
    U_{\text{model}} = \frac{9}{\sum_{m \in M} \frac{1}{m}}.
\end{equation}
\end{itemize}

\subsection{Models}
\label{Models}

Language models encode and store knowledge differently based on their architecture and training configuration, necessitating the evaluation of unlearning methods across diverse models to assess robustness and generalizability. However, existing benchmark implementations often support only a limited range of model types and require manual adaptation of evaluation logic, such as input formatting, tokenization, and prompting, when applied to new architectures.

\begin{table}[htp]
\centering
\begin{tabular}{cc}
\toprule
\textbf{Model} & \textbf{Params} \\
\midrule
Llama-2 & 7B, 7B-Chat \\
Llama-3.1 & 8B, 8B-Instruct \\
Llama-3.2 & 1B, 1B-Instruct, 3B, 3B-Instruct \\
\bottomrule
\end{tabular}
\caption{LLM architectures verified to run successfully within \obv.}
\label{tab:support_arch}
\end{table}

\obv addresses these challenges by supporting multiple model architectures and sizes natively. Built on Hugging Face Transformers~\cite{wolf2019huggingface}, it leverages \texttt{AutoModelForCausalLM} and \texttt{AutoTokenizer}, while also enabling custom model loading (e.g., for probe models). A unified abstraction facilitates seamless switching between chat-style and base models without modifying the unlearning or evaluation pipeline, reducing overhead and ensuring consistent cross-model comparisons. While Table~\ref{tab:support_arch} lists the models we have verified, the framework will support additional, more recent models as well.

\obv supports loading models in various precisions, including 4-bit and 8-bit quantized models via the \texttt{bitsandbytes} library~\cite{dettmers2022optimizers}. This quantization flexibility is particularly valuable for stress-testing unlearning methods, enabling robust evaluation across diverse computational constraints.

Additionally, \obv supports both full-parameter fine-tuning and parameter-efficient adaptation using Low-Rank Adaptation (LoRA)~\cite{hu2022lora}. This dual-mode fine-tuning capability enables flexible experimentation under varying resource constraints. In particular, LoRA-based fine-tuning allows users to efficiently adapt large models with significantly reduced memory and computational overhead, while still achieving competitive performance.




\subsection{Federated Methods}
\label{Federated}

Federated learning (FL) algorithms aim to train a robust global model by aggregating locally computed updates from distributed clients, with a strong emphasis on data privacy and system efficiency.

\subsubsection{FedAvg~\cite{mcmahan2017communication}:} 
    FedAvg is the foundational algorithm in FL. Each client independently trains a model on its local dataset and computes the update as the difference between the initial global model and the locally trained one. After local training, clients send their updates to a central server, which aggregates them via simple averaging to produce the new global model. All clients contribute equally to the aggregation, promoting fairness across heterogeneous data distributions. FedAvg reduces server load and preserves privacy by avoiding direct access to clients’ raw data.
    
    Federated Averaging computes a weighted aggregation of the clients' LoRA parameters at round~$t$:
\begin{equation}
    \phi_t = \sum_{k=1}^K \alpha_k \, \phi_k^{(t, R)}
\end{equation}
where $\phi_k^{(t, R)}$ denotes the LoRA parameters of client~$k$ after local training at round~$t$, and $\alpha_k$ is the aggregation weight for client~$k$, typically set to $\alpha_k = \frac{n_k}{\sum_{j=1}^K n_j}$, where $n_k$ is the number of local training examples at client~$k$.




\subsubsection{FedAvgM~\cite{hsu2019measuring}:} 
Based on the Federated Averaging algorithm, Federated Averaging with Momentum (FedAvgM) further incorporates server-side momentum to smooth model updates across communication rounds:
\begin{align}
    \Delta_t &= \sum_{k=1}^K \alpha_k \left( \phi_k^{(t, R)} - \phi_{t-1} \right) \label{eq:fedavgm_delta} \\
    m_t &= 
    \begin{cases}
        \Delta_t & \text{if } t = 0 \\
        \beta_1 m_{t-1} + \Delta_t & \text{if } t > 0
    \end{cases} \label{eq:fedavgm_momentum} \\
    \phi_t &= \phi_{t-1} + m_t \label{eq:fedavgm_update}
\end{align}

Here, $\phi_k^{(t, R)}$ represents the LoRA parameters obtained by client $k$ after local training in round~$t$, and $\phi_t$ denotes the global model at round~$t$. $\alpha_k$ is the aggregation weight, and $\beta_1 \in [0, 1)$ is the server momentum coefficient. The server maintains a velocity vector $m_t$ to accumulate update directions over rounds, thereby stabilizing the aggregation process and accelerating convergence.

\subsubsection{FedAdagrad~\cite{reddi2020adaptive}:}
Standard federated optimization methods, such as Federated Averaging (FedAvg), are often difficult to tune and exhibit unfavorable convergence behavior. In non-federated settings, adaptive optimization methods have had notable success in combating such issues. Therefore, Reddi~\etal propose federated versions of adaptive optimizers, including Adagrad, Adam, and Yogi, and analyze their convergence in the presence of heterogeneous data for general non-convex settings.

FedAdagrad is an instance of the FedOpt framework that applies Adagrad-style adaptive updates on the server while using SGD on the clients. The server update at communication round~$t$ is computed as:
\begin{align}
    \Delta_t &= \sum_{k=1}^K \alpha_k \left( \phi_k^{(t, R)} - \phi_{t-1} \right) \label{eq:fedadagrad_delta} \\
    v_t &= v_{t-1} + \Delta_t^2 \label{eq:fedadagrad_accum} \\
    \phi_t &= \phi_{t-1} + \epsilon \cdot \frac{\Delta_t}{\sqrt{v_t} + \tau} \label{eq:fedadagrad_update}
\end{align}

Here, $\phi_k^{(t, R)}$ denotes LoRA parameters of client~$k$ after local training at round~$t$, and $\phi_t$ is the global model at the server. $\Delta_t$ is the aggregated update, $v_t$ is the accumulated squared update, $\epsilon$ is the server learning rate, and $\tau > 0$ is a small constant that controls the degree of adaptivity and ensures numerical stability.

\subsubsection{FedAdam~\cite{reddi2020adaptive}:}
FedAdam adapts the Adam optimizer to the federated setting by maintaining server-side first and second moment estimates. The update rules at round~$t$ are as follows:
\begin{align}
    \Delta_t &= \sum_{k=1}^K \alpha_k \left( \phi_k^{(t, R)} - \phi_{t-1} \right) \label{eq:fedadam_delta} \\
    m_t &= \begin{cases}
        \Delta_t & \text{if } t = 0 \\
        \beta_1 m_{t-1} + (1 - \beta_1) \Delta_t & \text{otherwise}
    \end{cases} \label{eq:fedadam_momentum} \\
    v_t &= \beta_2 v_{t-1} + (1 - \beta_2) \Delta_t^2 \label{eq:fedadam_variance} \\
    \phi_t &= \phi_{t-1} + \epsilon \cdot \frac{m_t}{\sqrt{v_t} + \tau} \label{eq:fedadam_update}
\end{align}

Here, $\phi_k^{(t, R)}$ denotes the LoRA parameters of client~$k$ after local training in round~$t$, and $\phi_t$ is the global model. $\Delta_t$ is the aggregated update from clients, and $m_t$, $v_t$ are the first and second moment accumulators, respectively. The constants $\beta_1, \beta_2 \in [0, 1)$ are exponential decay rates, $\epsilon$ is the server learning rate, and $\tau > 0$ is a small constant to avoid division by zero and control adaptivity.

\subsubsection{FedYogi~\cite{reddi2020adaptive}:}
FedYogi modifies the update of the second moment by using a sign-based correction term to prevent the variance from growing too rapidly:
\begin{align}
    \Delta_t &= \sum_{k=1}^K \alpha_k (\phi_k^{(t,R)} - \phi_{t-1}) \\
    m_t &= \begin{cases} 
        \Delta_t & \text{if } t = 0 \\
        \beta_1 m_{t-1} + (1 - \beta_1) \Delta_t & \text{if } t > 0 
    \end{cases} \\
    v_t &= v_{t-1} - (1 - \beta_2)\Delta_t^2\operatorname{sign}(v_{t-1} - \Delta_t^2) \\
    \phi_t &= \phi_{t-1} + \epsilon \frac{m_t}{\sqrt{v_t} + \tau}
\end{align}

Here, $\phi_k^{(t, R)}$ denotes the LoRA parameters of client~$k$ after local training in round~$t$, and $\phi_t$ is the global model. $\Delta_t$ and $v_t$ are the first and second moment estimators on the server. The term $\operatorname{sign}(v_{t-1} - \Delta_t^2)$ ensures that $v_t$ does not increase monotonically, which improves stability over FedAdam.

\subsubsection{FedProx~\cite{li2020federated}:}
FedProx extends FedAvg by addressing challenges arising from data heterogeneity and partial client participation. It introduces a proximal term to the local training objective that penalizes divergence from the global model, thereby regularizing local updates. This modification constrains local models from straying too far from the global parameters, improving stability across rounds. The server aggregates the updates in a weighted manner, considering both data size and update consistency. As a result, FedProx yields a more stable and robust global model, especially in settings with non-IID data and intermittent client availability.

\begin{equation}
    \phi_t = \sum_{k=1}^K \alpha_k \phi_k^{(t,R)} \label{eq:fedprox_aggregation}
\end{equation}

And the client-side local objective includes a proximal term, as shown below:

\begin{equation}
    \mathcal{L}_k(\phi_k; \mathcal{D}_k) + \frac{\mu}{2} \left\| \phi_k - \phi_{t-1} \right\|^2, \label{eq:fedprox_local}
\end{equation}
where $\mu$ is the regularization coefficient. Aggregation remains identical to FedAvg.

\subsection{Unlearning Methods}
\label{Unlearning}

Unlearning methods constitute the foundation of our proposed \obv framework. However, in practice, researchers introducing novel unlearning techniques often restrict their evaluation to a single benchmark, primarily due to the substantial effort required to adapt implementations across diverse frameworks. This fragmentation has resulted in a notable absence of comprehensive, cross-benchmark comparative studies within the unlearning domain. The need to re-implement methods, reconcile differing evaluation protocols, and standardize metrics imposes significant overhead, thereby hindering reproducibility and impeding research progress.

\subsubsection{Retrain.}
Retraining involves re-optimizing an LLM model from scratch on a modified dataset that excludes specific data points or classes targeted for unlearning. This ensures the model no longer retains information about the removed data.
\subsubsection{Gradient Ascent~\cite{maini2024tofu}.} This technique performs gradient ascent on the data designated for forgetting, aiming to systematically diminish the model’s confidence in the targeted examples, thereby facilitating effective unlearning.

    \begin{equation}\mathcal{L}=-\gamma\mathbb{E}_{(x,y_\mathrm{f})\thicksim\mathcal{D}_\mathrm{forget}}\ell\left(y_\mathrm{f}|x;f_\mathrm{unl}\right)\end{equation}

\subsubsection{GradDiff~\cite{maini2024tofu}.} This method optimizes the model by performing gradient ascent on the forget data to degrade confidence on targeted samples, concurrently applying gradient descent on the retain data to maintain performance on the remaining data.
    \begin{equation}\mathcal{L}=-\gamma\mathbb{E}_{(x,y_\mathrm{f})\thicksim\mathcal{D}_\mathrm{forget}}\ell\left(y_\mathrm{f}|x;f_\mathrm{unl}\right)+\alpha\mathbb{E}_{(x,y)\thicksim\mathcal{D}_\mathrm{retain}}\ell\left(y|x;f_\mathrm{unl}\right)\end{equation}

\subsubsection{NPO~\cite{zhang2024negative}.} Similar in spirit to the DPO-style objective, NPO utilizes only the negative feedback component during optimization. This selective focus leads to enhanced training stability relative to comparable methods like GradDiff.
    \begin{equation}
    \begin{aligned}
    \mathcal{L} &= -\frac{2}{\beta} \mathbb{E}_{(x, y_\mathrm{f}) \sim \mathcal{D}_\mathrm{forget}} \log \sigma \left( -\beta \log \left( \frac{p(y_\mathrm{f} | x; f_\mathrm{unl})}{p(y_\mathrm{f} | x; f_\mathrm{target})} \right) \right) \\
    &\quad + \alpha \mathbb{E}_{(x, y) \sim \mathcal{D}_\mathrm{retain}} \ell \left( y | x; f_\mathrm{unl} \right)
    \end{aligned}
    \end{equation}

\subsubsection{SimNPO~\cite{fan2024simplicity}.} This method is a modified variant of NPO, which maintains the core forgetting behavior by replacing the reference model with a delta ($\delta$) term within the loss function, enhancing flexibility in the optimization process.
    \begin{equation}
    \begin{aligned}
    \mathcal{L} &= -\frac{2}{\beta} \mathbb{E}_{(x,y_\mathrm{f}) \sim \mathcal{D}_\mathrm{forget}} \log \sigma \left( -\frac{\beta}{|y_\mathrm{f}|} \log p(y_\mathrm{f} | x; f_\mathrm{unl}) - \delta \right) \\
    &\quad + \alpha \mathbb{E}_{(x,y) \sim \mathcal{D}_\mathrm{retain}} \ell \left( y | x; f_\mathrm{unl} \right)
    \end{aligned}
    \end{equation}


\section{Experimental Details}
\label{Experimental}
\subsection{Testbed}
\label{Testbed}
\subsubsection{Hardware Configuration.} All experiments are conducted on an Elastic Compute Service (ECS) instance equipped with an Intel(R) Xeon(R) Platinum 8369B CPU (32 cores available), 256 GB of RAM, 512 GB of free disk space, and an NVIDIA A100 GPU with 80 GB of memory. For reproducibility and consistent performance, we recommend using the \texttt{ecs.gn7e-c16g1.8xlarge} instance type on Alibaba Cloud.

\subsubsection{Software Environment.} We conduct all experiments on Ubuntu 22.04.5 LTS with NVIDIA driver version 535.216.03 and CUDA 12.2. The implementation of the \obv framework is based on Python 3.12.11 and PyTorch 2.7.1. A complete list of dependencies is provided in the accompanying \texttt{requirements.txt}, and additional installation and usage instructions are included in the \texttt{README.md} file within the submitted code.

\subsection{Hyperparameters}
\label{Hyperparameters}

We conduct experiments using 30 clients with a $10\%$ participation rate for \obv. In each round, a randomly selected client requests targeted sample-level unlearning. The training process consists of 5 local epochs and 10 global rounds, with a one-epoch warmup period included on four base models: \textit{Llama-2-7b-hf}, \textit{Llama-3.1-8B-Instruct}, \textit{Llama-3.2-1B-Instruct}, and \textit{Llama-3.2-3B-Instruct}. We also adopt the Low-Rank Adaptation (LoRA) technique to fine-tune our base model. The LoRA modules are injected into key components of the transformer architecture, including \{\texttt{q\_proj}, \texttt{k\_proj}, \texttt{v\_proj}, \texttt{o\_proj}, \texttt{gate\_proj}, \texttt{up\_proj}, \texttt{down\_proj}\}.
For more details, the key hyperparameters used in our experiments across different domains are summarized in Table~\ref{tab:fl_hyperparams}.

\subsection{Prompt Template}
\label{Prompt}
\begin{itemize}
    \item \textbf{Llama-2}: The chat template is disabled during training. User queries and assistant responses follow the format, as shown in Figure~\ref{pmp2}.

\item \textbf{Llama-3.1 \& 3.2}: The chat template is enabled during training. User queries and assistant responses follow the format, as illustrated in Figure~\ref{pmp3}.

\end{itemize}

\subsection{Overheads}
\label{Overheads}
Table~\ref{table:cost} reports the computational costs of various FL algorithms across three stages: \textbf{Retain}, \textbf{Finetune}, and \textbf{Unlearn}. Among the five FL algorithms evaluated, specifically FedAvg, FedAvgM, FedAdagrad, FedAdam, and FedYogi, runtimes are comparable, with each stage completing in approximately 700 to 760 seconds. In contrast, FedProx incurs significantly higher costs in all phases, exceeding 1600 seconds in both Retain and Finetune, and around 1570 seconds across unlearning variants. This increase is primarily attributed to its proximal regularization term, which adds computational burden during local updates. Among the unlearning methods, runtime differences are marginal, with NPO and SimNPO slightly higher than GradAscent and GradDiff.

Overall, the choice of FL algorithm remains the dominant factor affecting total computational cost.
\section{Limitations}
\label{Limitations}

While \obv demonstrates robust performance for federated LLM unlearning across the TOFU and MUSE datasets, it exhibits limitations on the MUSE Books set, as shown in Table~\ref{suptable3}, where models show suboptimal performance in long-context and few-shot scenarios. This is likely due to information sparsity in extended sequences and limited training samples hindering generalization. Additionally, the computational cost of joint optimization in \obv may challenge resource-constrained environments. These insights guide future research toward enhancing long-context processing with context-aware attention mechanisms and improving few-shot learning via data augmentation or meta-learning, alongside optimizing communication efficiency for scalable FedLLM deployments.

\section{Supplementary Experiments}
\label{Supplementary Experiments}
\begin{figure*}[htp]
    \centering
    \includegraphics[width=1.0\linewidth]{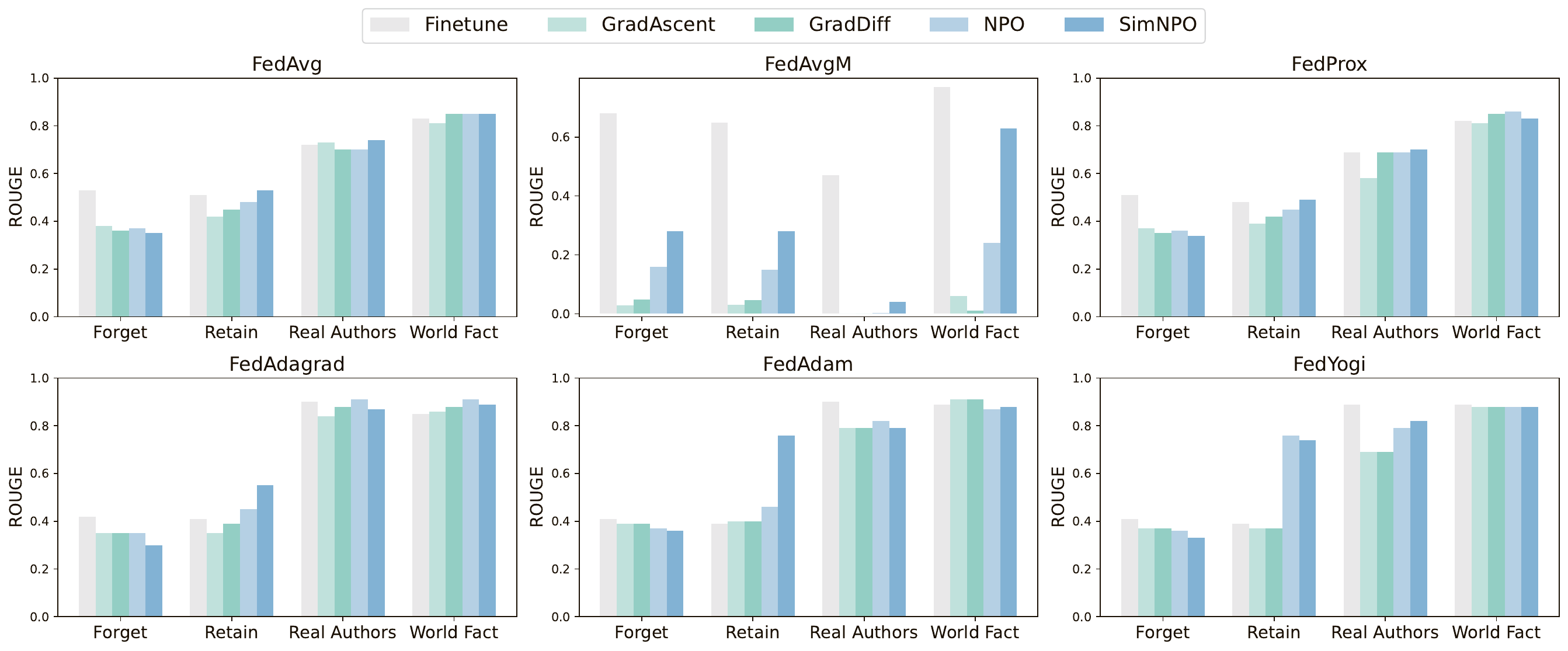}
    \caption{Comparative analysis of \textbf{ROUGE} scores across federated learning and unlearning methods using \textbf{Llama-3.2-3B} model with \textbf{Split90} strategies. For the Forget set, lower scores indicate better performance ($\downarrow$), whereas for the remaining sets, higher scores are preferable ($\uparrow$).}
    
    \label{supexp1}
\end{figure*}

\begin{figure*}[t]
    \centering
    \includegraphics[width=1.0\linewidth]{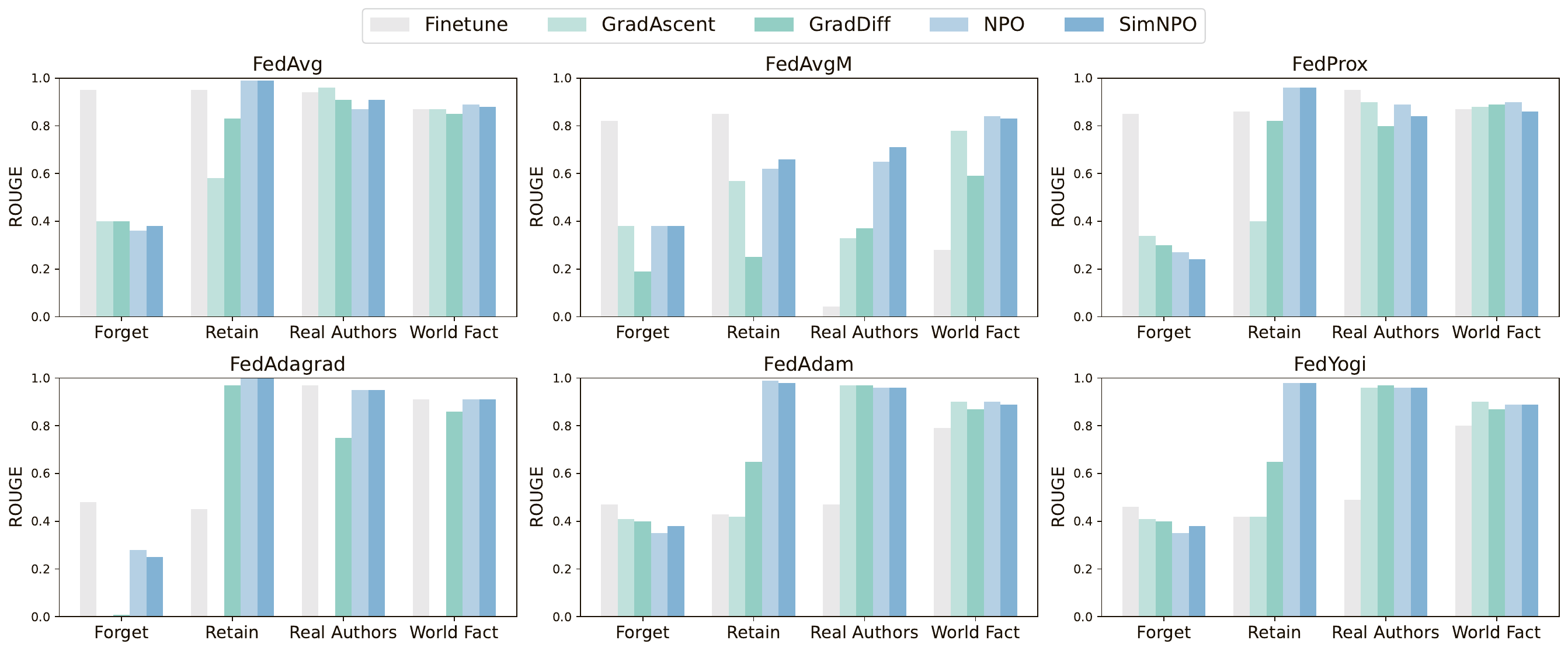}
    \caption{Comparative analysis of \textbf{ROUGE} scores across federated learning and unlearning methods using \textbf{Llama-3.1-8B} model with \textbf{Split90} strategies. For the Forget set, lower scores indicate better performance ($\downarrow$), whereas for the remaining sets, higher scores are preferable ($\uparrow$).}
    \label{supexp8b_R_90}
\end{figure*}

\begin{figure*}[t]
    \centering
    \includegraphics[width=1.0\linewidth]{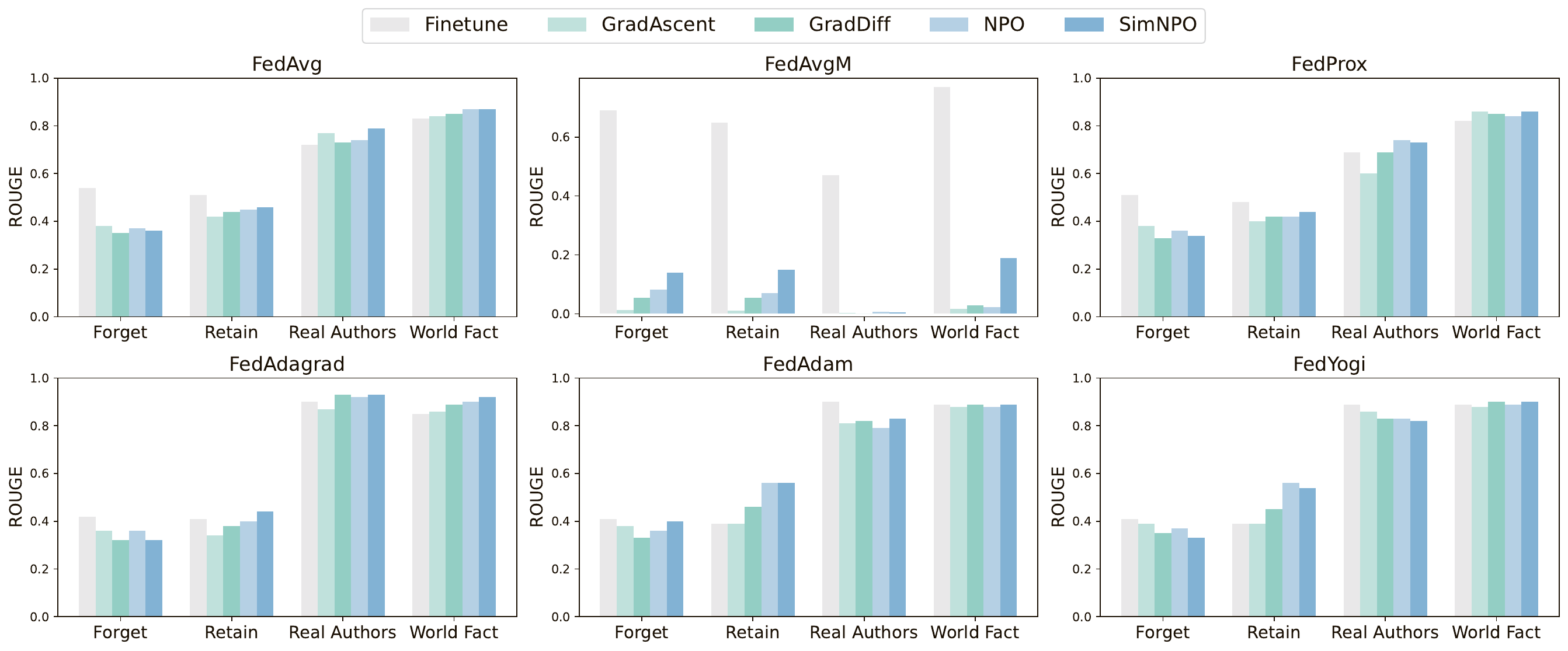}
    \caption{Comparative analysis of \textbf{ROUGE} scores across federated learning and unlearning methods using \textbf{Llama-3.2-3B} model with \textbf{Split95} strategies. For the Forget set, lower scores indicate better performance ($\downarrow$), whereas for the remaining sets, higher scores are preferable ($\uparrow$).}
    \label{supexp2}
\end{figure*}

\begin{figure*}[t]
    \centering
    \includegraphics[width=1.0\linewidth]{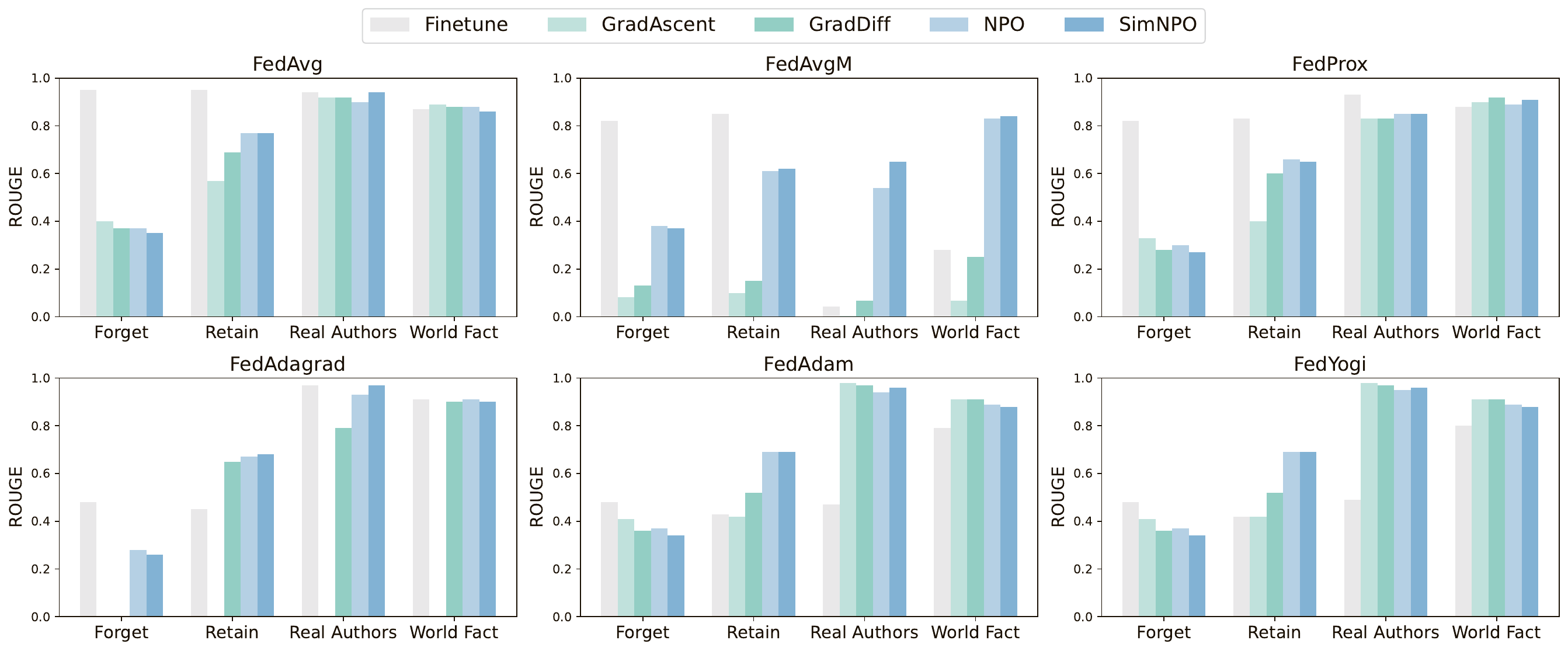}
    \caption{Comparative analysis of \textbf{ROUGE} scores across federated learning and unlearning methods using \textbf{Llama-3.1-8B} model with \textbf{Split95} strategies. For the Forget set, lower scores indicate better performance ($\downarrow$), whereas for the remaining sets, higher scores are preferable ($\uparrow$).}
    \label{supexp8b_R_95}
\end{figure*}

\begin{figure*}[t]
    \centering
    \includegraphics[width=1.0\linewidth]{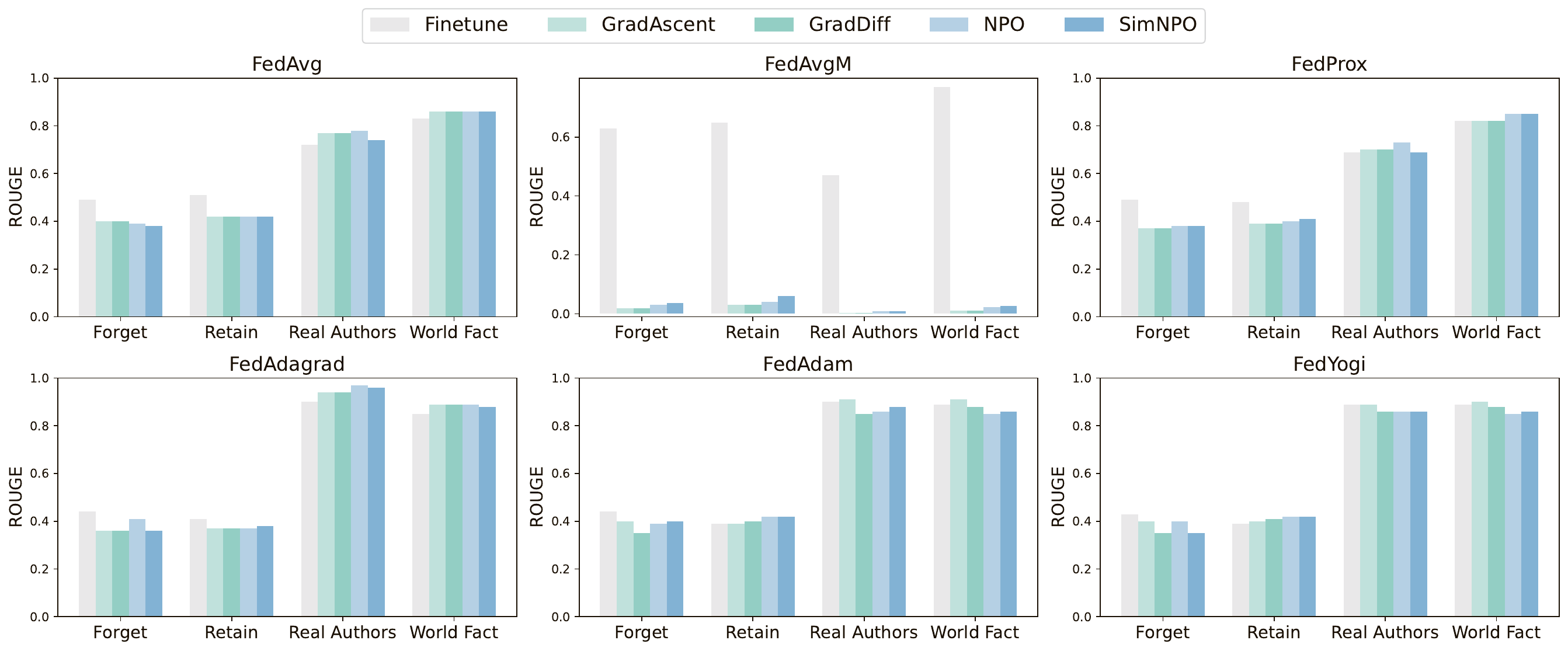}
    \caption{Comparative analysis of \textbf{ROUGE} scores across federated learning and unlearning methods using \textbf{Llama-3.2-3B} model with \textbf{Split99} strategies. For the Forget set, lower scores indicate better performance ($\downarrow$), whereas for the remaining sets, higher scores are preferable ($\uparrow$).}
    \label{supexp3}
\end{figure*}

\begin{figure*}[t]
    \centering
    \includegraphics[width=1.0\linewidth]{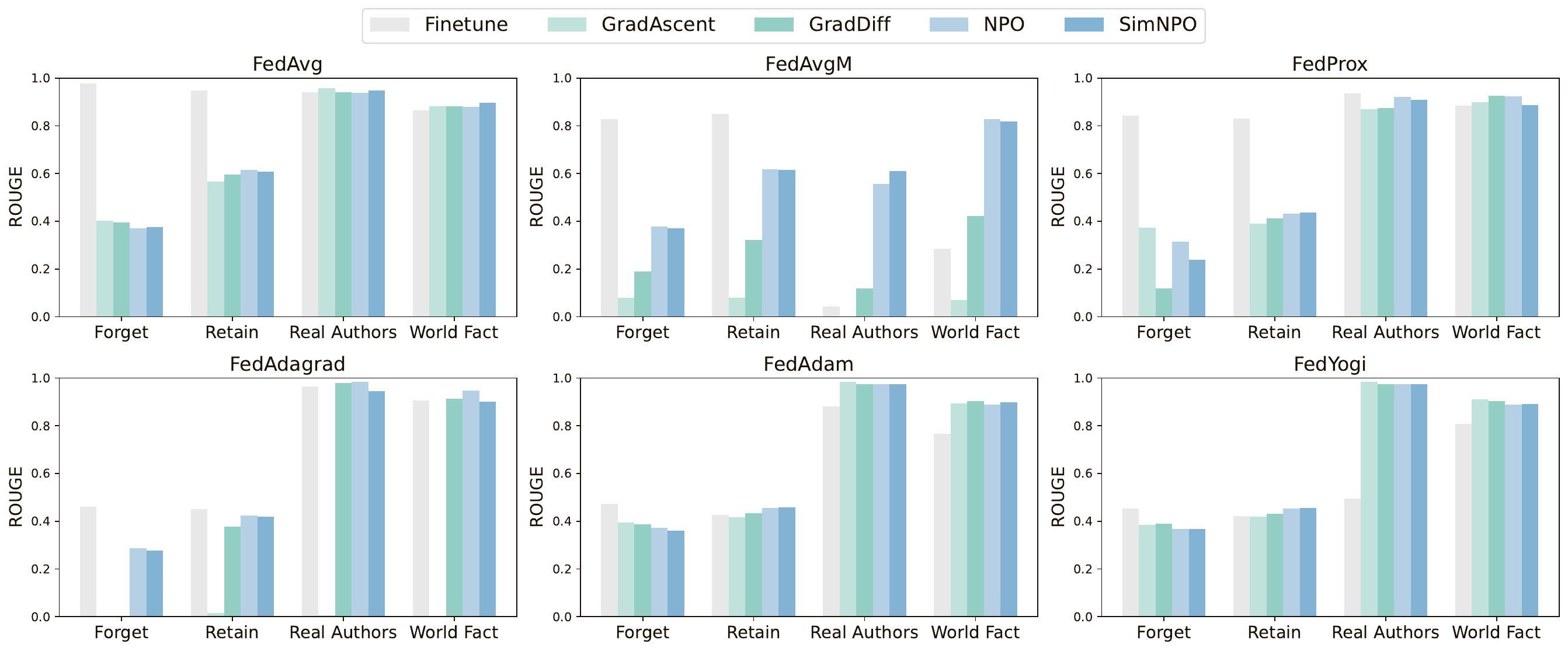}
    \caption{Comparative analysis of \textbf{ROUGE} scores across federated learning and unlearning methods using \textbf{Llama-3.1-8B} model with \textbf{Split99} strategies. For the Forget set, lower scores indicate better performance ($\downarrow$), whereas for the remaining sets, higher scores are preferable ($\uparrow$).}
    \label{supexp8b_R_99}
\end{figure*}

\begin{figure*}[t]
    \centering
    \includegraphics[width=1.0\linewidth]{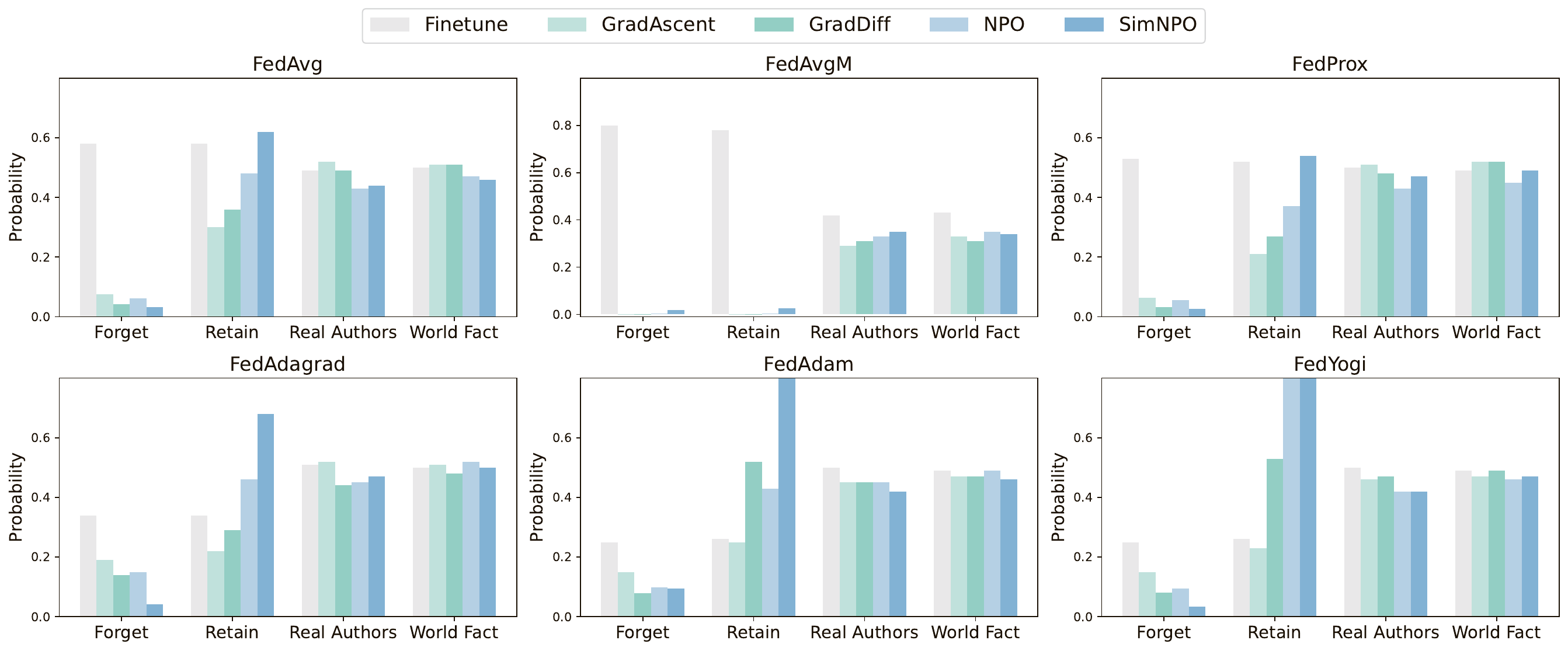}
    \caption{Comparative analysis of \textbf{Probability} scores across federated learning and unlearning methods using \textbf{Llama-3.2-3B} model with \textbf{Split90} strategies. For the Forget set, lower scores indicate better performance ($\downarrow$), whereas for the remaining sets, higher scores are preferable ($\uparrow$).}
    \label{supexp4}
\end{figure*}

\begin{figure*}[t]
    \centering
    \includegraphics[width=1.0\linewidth]{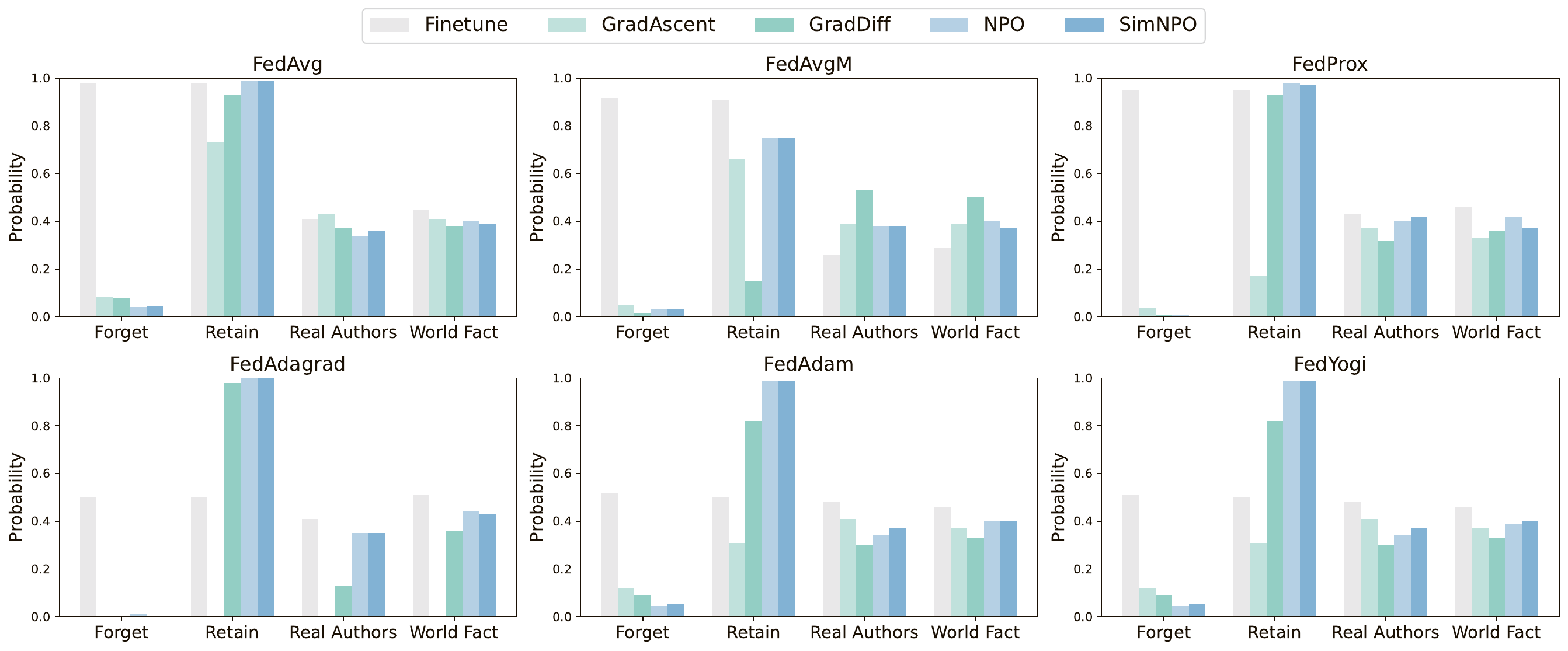}
    \caption{Comparative analysis of \textbf{Probability} scores across federated learning and unlearning methods using \textbf{Llama-3.1-8B} model with \textbf{Split90} strategies. For the Forget set, lower scores indicate better performance ($\downarrow$), whereas for the remaining sets, higher scores are preferable ($\uparrow$).}
    \label{supexp8b_P_90}
\end{figure*}

\begin{figure*}[t]
    \centering
    \includegraphics[width=1.0\linewidth]{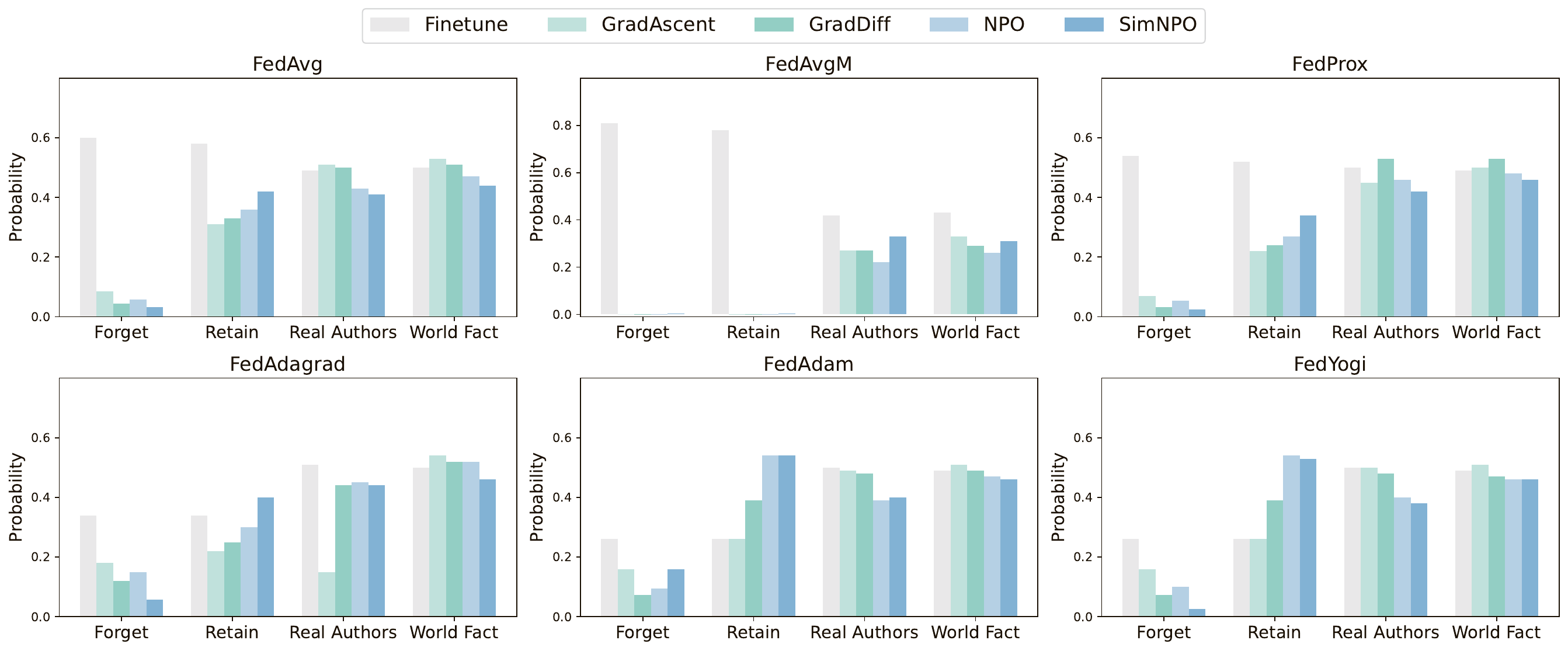}
    \caption{Comparative analysis of \textbf{Probability} scores across federated learning and unlearning methods using \textbf{Llama-3.2-3B} model with \textbf{Split95} strategies. For the Forget set, lower scores indicate better performance ($\downarrow$), whereas for the remaining sets, higher scores are preferable ($\uparrow$).}
    \label{supexp5}
\end{figure*}

\begin{figure*}[t]
    \centering
    \includegraphics[width=1.0\linewidth]{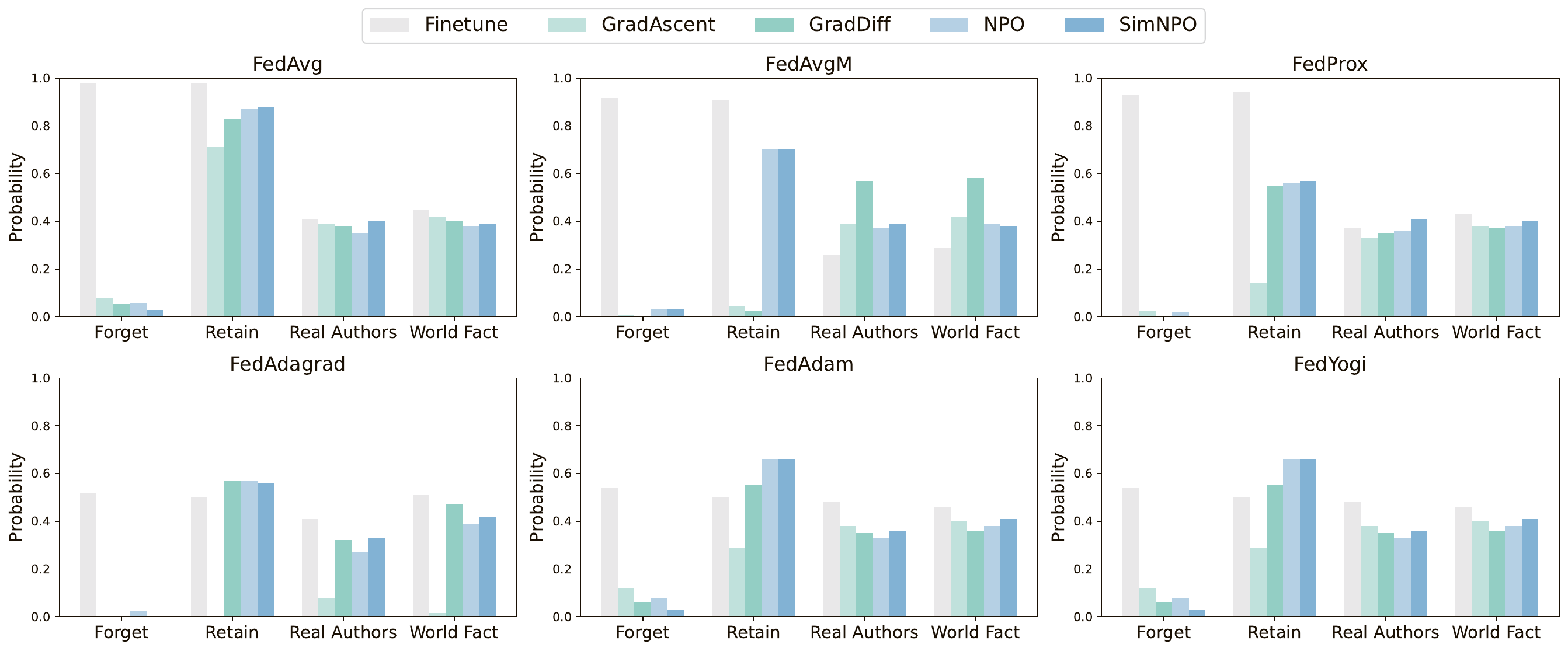}
    \caption{Comparative analysis of \textbf{Probability} scores across federated learning and unlearning methods using \textbf{Llama-3.1-8B} model with \textbf{Split95} strategies. For the Forget set, lower scores indicate better performance ($\downarrow$), whereas for the remaining sets, higher scores are preferable ($\uparrow$).}
    \label{supexp8b_P_95}
\end{figure*}

\begin{figure*}[t]
    \centering
    \includegraphics[width=1.0\linewidth]{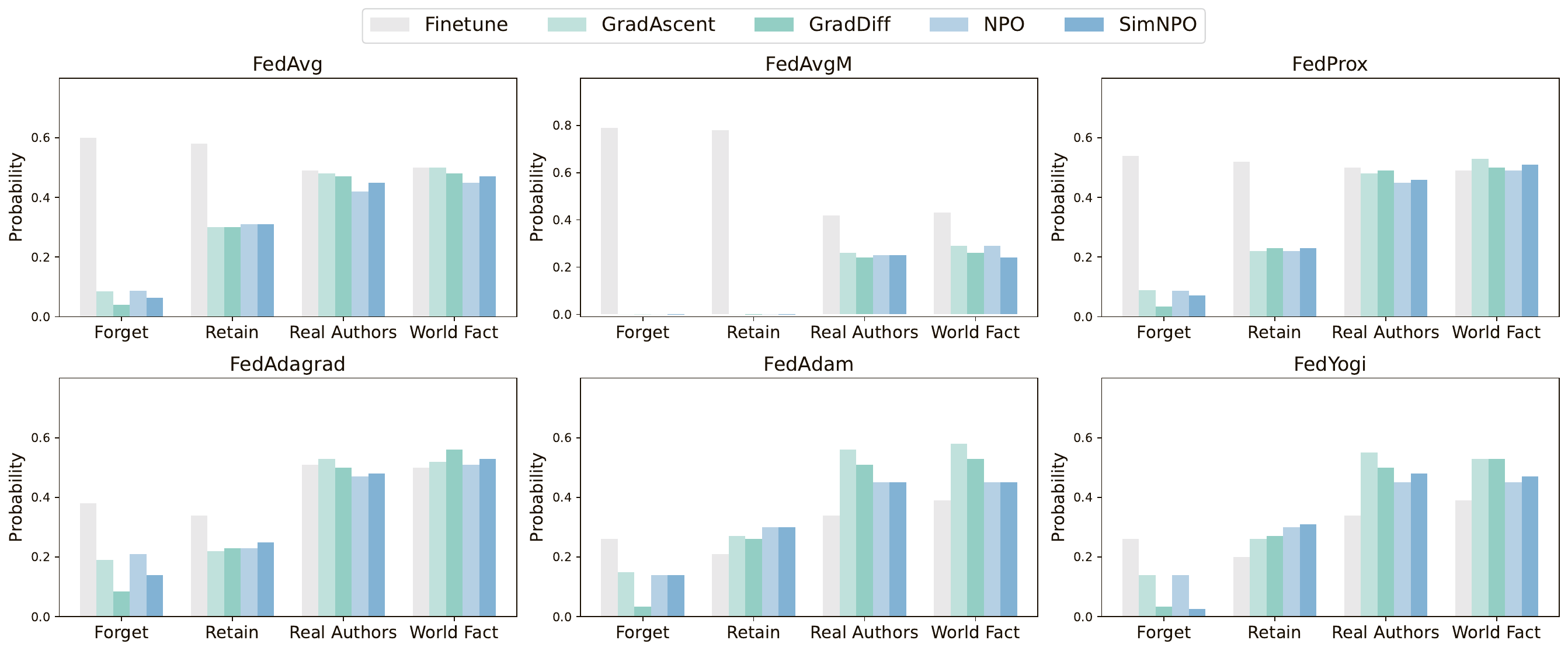}
    \caption{Comparative analysis of \textbf{Probability} scores across federated learning and unlearning methods using \textbf{Llama-3.2-3B} model with \textbf{Split99} strategies. For the Forget set, lower scores indicate better performance ($\downarrow$), whereas for the remaining sets, higher scores are preferable ($\uparrow$).}
    \label{supexp6}
\end{figure*}

\begin{figure*}[t]
    \centering
    \includegraphics[width=1.0\linewidth]{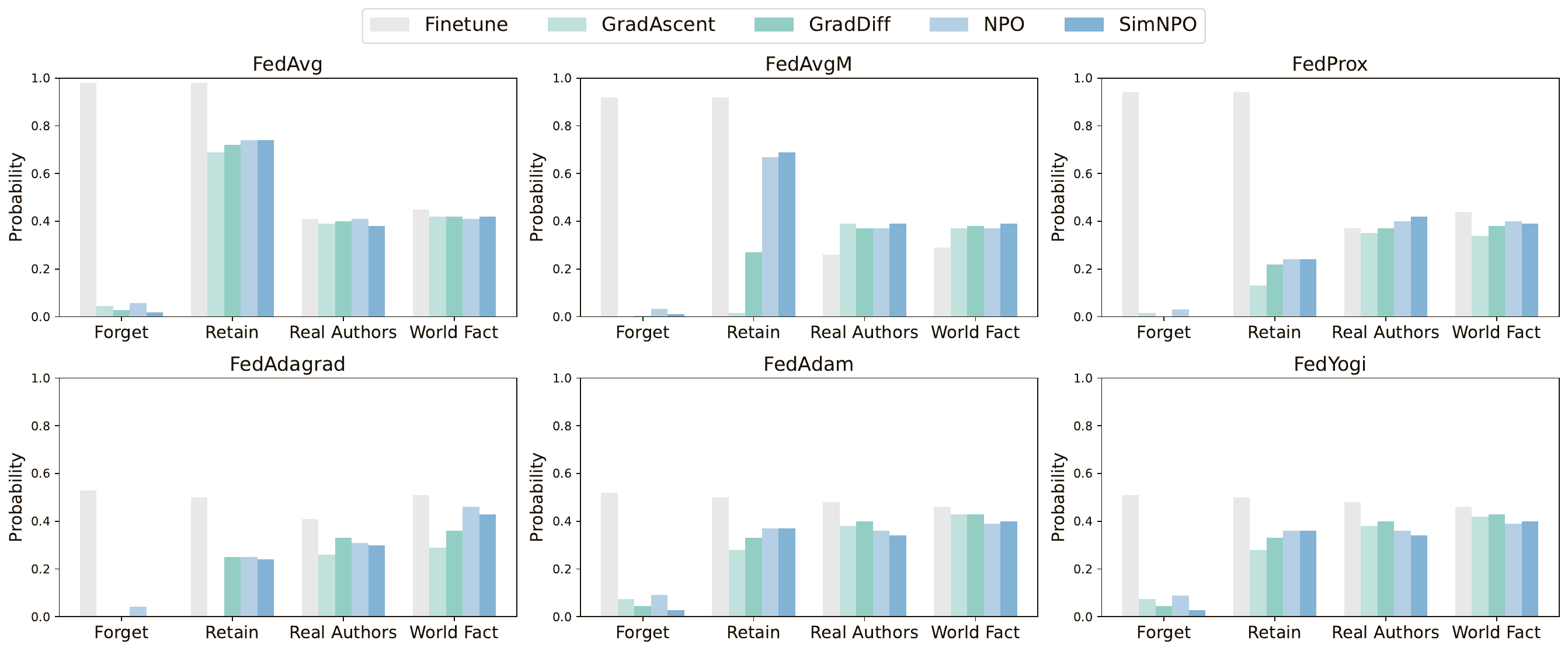}
    \caption{Comparative analysis of \textbf{Probability} scores across federated learning and unlearning methods using \textbf{Llama-3.1-8B} model with \textbf{Split99} strategies. For the Forget set, lower scores indicate better performance ($\downarrow$), whereas for the remaining sets, higher scores are preferable ($\uparrow$).}
    \label{supexp8b_P_99}
\end{figure*}

In this section, we conduct extensive supplementary experiments to rigorously evaluate the robustness and scalability of \obv for Federated Large Language Models (FedLLMs) and Federated LLM Unlearning (FedLLMU). Our evaluation spans a range of model sizes and data splits, with detailed results presented in Figures~\ref{supexp1}–\ref{supexp8b_P_99} and Tables~\ref{suptable2}–\ref{suptable3}. These findings demonstrate the framework’s versatility and effectiveness across diverse privacy-sensitive settings.

Additionally, Figure~\ref{fig:tofu_qa_examples_2x2} illustrates a comparative analysis of model responses across various datasets, underscoring \obv’s effectiveness in targeted unlearning tasks.


\begin{table*}[!htp]
    \centering
    \begin{tabular}{cccc}
        \toprule
        \textbf{Notation} & \textbf{Hyperparameter} & \textbf{Value} & \textbf{Explanation} \\
        \midrule
        \rowcolor{cyan!5}
        \multicolumn{4}{c}{\textbf{General Training Configuration}} \\
        \midrule
        -- & Seed & 0 & Seed Value for Randomness Control \\
        -- & Precision & \texttt{bfloat16} & Numeric Format Used in Training \\
        -- & Attention Mechanism & Flash Attention 2 & Optimized Attention Computation \\
        -- & Optimizer & \texttt{paged\_adamw\_32bit} & Memory-Efficient AdamW Variant \\
        $\eta$ & Learning Rate & $8 \times 10^{-5}$ & Initial Learning Rate \\
        $\lambda_{\text{wd}}$ & Weight Decay & 0.01 & L2 Regularization Strength \\
        $B$ & Batch Size & 32 & Number of Samples per Batch \\
        \midrule
        \rowcolor{cyan!5}
        \multicolumn{4}{c}{\textbf{Federated Optimization Settings}} \\
        \midrule
        $\eta_s$ & Server Learning Rate & 1.0 & Learning Rate for Global Model Update \\
        $\beta_1$ & First Moment Decay Rate & 0.9 & Exponential Decay for Mean Estimate \\
        $\beta_2$ & Second Moment Decay Rate & 0.99 & Exponential Decay for Variance Estimate \\
        $\epsilon$ & Epsilon for Stability & $1 \times 10^{-3}$ & Numerical Stability Constant \\
        $\lambda$ & Regularization Term & $1 \times 10^{-3}$ & Regularization Weight on Global Model \\
        $\mu$ & Proximal Coefficient & 0.01 & Strength of FedProx Proximal Term \\
        $\gamma$ & Momentum Factor & 0.9 & Momentum in Server Update \\
        \midrule
        \rowcolor{cyan!5}
        \multicolumn{4}{c}{\textbf{LoRA Configuration}} \\
        \midrule
        $r$ & Rank & 32 & Rank of LoRA Decomposition \\
        $\alpha$ & LoRA Scaling & 64 & Scaling Factor for LoRA Layers \\
        $p$ & Dropout & 0.05 & Dropout Rate in LoRA Modules \\
        $b$ & Bias & None & Bias Strategy in LoRA (None, All, or LoRA-Only) \\
        \bottomrule
    \end{tabular}
    \caption{Default configurations and chosen hyperparameters adopted in our experimental setup.}
    \label{tab:fl_hyperparams}
\end{table*}

\begin{figure*}[!htp]
\centering
\begin{minipage}{\textwidth}
\begin{pmt}{Llama-2 Series Template}
\textbf{Question}: [Content]\textbackslash n\\
\textbf{Answer}: [Content]\textbackslash n\textbackslash n
\end{pmt}
\end{minipage}
\caption{Prompt template for Llama-2 series.}
\label{pmp2}
\end{figure*}

\begin{figure*}[!htp]
\centering
\begin{minipage}{\textwidth}
\begin{pmt}{Llama-3.1 \& 3.2 Series Template}
\textbf{System Prompt}: You are a helpful assistant.\\
\textbf{System Prompt with Special Tokens}: \texttt{<|begin\_of\_text|><|start\_header\_id|>}system\texttt{<|end\_header\_id|>}\\\texttt{\textbackslash n\textbackslash n}You are a helpful assistant.\texttt{<|eot\_id|>}\\
\textbf{User Start Tag}: \texttt{<|start\_header\_id|>}user\texttt{<|end\_header\_id|>}\\
\textbf{User End Tag}: \texttt{<|eot\_id|>}\\
\textbf{Asst Start Tag}: \texttt{<|start\_header\_id|>}assistant\texttt{<|end\_header\_id|>}\\
\textbf{Asst End Tag}: \texttt{<|eot\_id|>}\\
\textbf{Data String}: \texttt{10 Apr 2025}
\end{pmt}
\end{minipage}
\caption{Prompt template for Llama-3.1 \& 3.2 series.}
\label{pmp3}
\end{figure*}

\begin{table*}[!htp]
  \centering
  \begin{tabular}{ccccccc}
    \toprule
    \multirow{2}[2]{*}{\textbf{Algorithms}} & \multirow{2}[2]{*}{\textbf{Retain (s)}} & \multirow{2}[2]{*}{\textbf{Finetune (s)}} & \multicolumn{4}{c}{\textbf{Unlearn (s)}} \\
    \cmidrule(lr){4-7}
    & & & \textbf{GradAscent} & \textbf{GradDiff} & \textbf{NPO} & \textbf{SimNPO} \\
    \midrule
    FedAvg     & 752.19 & 755.08 & 696.16 & 698.93 & 710.22 & 705.88 \\
    FedAvgM    & 758.98 & 754.74 & 696.72 & 700.07 & 714.39 & 709.33 \\
    FedAdagrad & 761.46 & 750.65 & 695.73 & 699.57 & 712.15 & 704.52 \\
    FedAdam    & 749.56 & 748.83 & 694.22 & 698.39 & 711.03 & 705.00 \\
    FedYogi    & 757.84 & 748.19 & 696.54 & 699.66 & 710.69 & 705.79 \\
    FedProx    & 1629.37 & 1612.86 & 1577.00 & 1565.71 & 1575.29 & 1567.06 \\
    \bottomrule
  \end{tabular}
  \caption{Comparison of computational costs across FL methods on the \textbf{TOFU} dataset, using the \textbf{Llama-3.2-1B} model with \textbf{Split99} strategy. The unlearning phase considers four variants: GradAscent, GradDiff, NPO, and SimNPO.}
  \label{table:cost}
\end{table*}

\begin{table*}[htp]
  \centering
  \begin{tabular}{ccccccccccccc}
  \toprule
    \multirow{3}[3]{*}{\textbf{Algorithms}} & \multicolumn{6}{c}{\textbf{Weighted Averaging-Based FL}} & \multicolumn{6}{c}{\textbf{Adaptive Optimization FL}} \\
     \cmidrule(lr){2-7} \cmidrule(lr){8-13}
     & \multicolumn{2}{c}{\textbf{FedAvg}} & \multicolumn{2}{c}{\textbf{FedAvgM}} & \multicolumn{2}{c}{\textbf{FedProx}} & \multicolumn{2}{c}{\textbf{FedAdagrad}} & \multicolumn{2}{c}{\textbf{FedAdam}} & \multicolumn{2}{c}{\textbf{FedYogi}} \\
     \cmidrule(lr){2-3} \cmidrule(lr){4-5} \cmidrule(lr){6-7} \cmidrule(lr){8-9} \cmidrule(lr){10-11} \cmidrule(lr){12-13}
     & MU$\uparrow$ & FTR$\uparrow$ & MU$\uparrow$ & FTR$\uparrow$ & MU$\uparrow$ & FTR$\uparrow$ & MU$\uparrow$ & FTR$\uparrow$ & MU$\uparrow$ & FTR$\uparrow$ & MU$\uparrow$ & FTR$\uparrow$\\
    \toprule
    \rowcolor{cyan!5}
    \multicolumn{13}{c}{\textbf{Meta Llama-3.2-1B-Instruct with LoRA}} \\
    \midrule
    \rowcolor{gray!10} \textbf{Finetune} & 0.50 & 0.50 & 0.48 & 0.42 & 0.50 & 0.51 & 0.45 & 0.65 &0.47 &0.55&0.47&0.53 \\
    \textbf{GradAscent} & \cellcolor{cyan!20}{\textbf{0.45}} & 0.59 & 0.00030 & 0.68 & 0.43 & 0.59 & 0.34 & \underline{0.71} & 0.44 & 0.64 & 0.44 & 0.64 \\
    \textbf{GradDiff} & \cellcolor{cyan!20}{\textbf{0.46}} & 0.58 & 0.00095 & \underline{0.69} & 0.43 & 0.57 & 0.43 & 0.33 & 0.45 & 0.63 & 0.45 & 0.63 \\
    \textbf{NPO} & 0.47 & 0.59 & 0.0030 & 0.64 & 0.46 & 0.58 & \cellcolor{cyan!20}{\textbf{0.49}} & \underline{0.68} & \cellcolor{cyan!20}{\textbf{0.49}} & 0.63 & \cellcolor{cyan!20}{\textbf{0.49}} & 0.63 \\
    \textbf{SimNPO} & 0.49 & 0.58 & 0.043 & 0.59 & 0.48 & 0.58 & \cellcolor{cyan!20}{\textbf{0.54}} & \underline{0.61} & 0.51 & \underline{0.61} & 0.51 & \underline{0.61} \\
    \textbf{Retrain} & \cellcolor{cyan!20}{\textbf{0.52}} & 0.63 & 0.47 & 0.58 & 0.51 & 0.64 & 0.46 & 0.69 & 0.32 & \underline{0.72} & 0.32 & \underline{0.72} \\
    \midrule
    \rowcolor{cyan!5}
    \multicolumn{13}{c}{\textbf{Meta Llama-3.2-3B-Instruct with LoRA}} \\
    \midrule
    \rowcolor{gray!10} \textbf{Finetune} & \cellcolor{cyan!20}{\textbf{0.59}} & 0.47 & 0.56 & 0.44 & 0.58 & 0.49 & 0.53 & \underline{0.62} & 0.50 & 0.58 & 0.5 & 0.58 \\
    \textbf{GradAscent} & \cellcolor{cyan!20}{\textbf{0.52}} & 0.58 & 0 & \underline{0.75} & 0.47 & 0.57 & 0.44 & 0.70 & 0.48 & 0.63 & 0.46 & 0.63 \\
    \textbf{GradDiff} & 0.54 & 0.55 & 0 & \underline{0.71} & 0.51 & 0.52 & 0.48 & 0.68 & 0.57 & 0.61 & \cellcolor{cyan!20}{\textbf{0.58}} & 0.62 \\
    \textbf{NPO} & 0.55 & 0.58 & 0.014 & 0.63 & 0.52 & 0.58 & 0.56 & \underline{0.67} & 0.55 & 0.62 & \cellcolor{cyan!20}{\textbf{0.61}} & 0.63 \\
    \textbf{SimNPO} & 0.58 & 0.59 & 0.11 & 0.58 & 0.58 & 0.58 & \cellcolor{cyan!20}{\textbf{0.62}} & \underline{0.63} & \cellcolor{cyan!20}{\textbf{0.62}} & \underline{0.63} & \cellcolor{cyan!20}{\textbf{0.62}} & 0.60 \\
    \textbf{Retrain} & \cellcolor{cyan!20}{\textbf{0.58}} & 0.64 & \cellcolor{cyan!20}{\textbf{0.58}} & 0.59 & 0.57 & 0.64 & 0.52 & \underline{0.68} & 0.51 & 0.63 & 0.51 & 0.63 \\
    \midrule
    \rowcolor{cyan!5}
    \multicolumn{13}{c}{\textbf{Meta Llama-3.1-8B-Instruct with LoRA}} \\
    \midrule
    \rowcolor{gray!10} \textbf{Finetune} & \cellcolor{cyan!20}{\textbf{0.64}} & 0.45 & 0.21 & 0.45 & \cellcolor{cyan!20}{\textbf{0.64}} & 0.44 & 0.55 & \underline{0.53} & 0.53 & 0.45 & 0.54 & 0.45 \\
    \textbf{GradAscent} & \cellcolor{cyan!20}{\textbf{0.59}} & 0.59 & 0.49 & 0.58 & 0.4 & 0.55 & 0 & 1.9e-21 & 0.48 & \underline{0.62} & 0.48 & \underline{0.62} \\
    \textbf{GradDiff} & \cellcolor{cyan!20}{\textbf{0.57}} & 0.58 & 0.38 & 0.5 & 0.54 & 0.48 & 0.43 & 3.3e-10 & 0.51 & \underline{0.59} & 0.51 & \underline{0.59} \\
    \textbf{NPO} & 0.58 & 0.57 & 0.55 & 0.52 & \cellcolor{cyan!20}{\textbf{0.61}} & 0.58 & 0.59 & \underline{0.59} & 0.58 & \underline{0.59} & 0.58 & \underline{0.59} \\
    \textbf{SimNPO} & 0.58 & \underline{0.60} & 0.55 & 0.54 & \cellcolor{cyan!20}{\textbf{0.60}} & 0.48 & \cellcolor{cyan!20}{\textbf{0.60}} & 0.49 & \cellcolor{cyan!20}{\textbf{0.60}} & \underline{0.60} & 0.59 & \underline{0.60} \\
    \textbf{Retrain} & \cellcolor{cyan!20}{\textbf{0.62}} & 0.62 & 0.20 & 0.61 & 0.61 & 0.60 & 0.53 & \underline{0.66} & 0.51 & 0.59 & 0.52 & 0.59 \\
    \bottomrule
  \end{tabular}
  \caption{Performance comparison of federated learning and unlearning algorithms on the \textbf{TOFU} dataset using \textbf{Llama-3.1-8B, Llama-3.2-1B and 3B} models, evaluated on metrics MU (Model Utility) and FTR (Forget Truth Ratio) with \textbf{Split90} strategies. Scores in \cellcolor{cyan!20}{\textbf{Bold}} indicate the optimal MU in different FL methods, while scores \underline{underlined} indicate the optimal FTR in different FL methods.}
  \label{suptable2}
\end{table*}

\begin{table*}[t]
  \centering
  \begin{tabular}{ccccccccccccc}
  \toprule
    \multirow{3}[3]{*}{\textbf{Algorithms}} & \multicolumn{6}{c}{\textbf{Weighted Averaging-Based FL}} & \multicolumn{6}{c}{\textbf{Adaptive Optimization FL}} \\
     \cmidrule(lr){2-7} \cmidrule(lr){8-13}
     & \multicolumn{2}{c}{\textbf{FedAvg}} & \multicolumn{2}{c}{\textbf{FedAvgM}} & \multicolumn{2}{c}{\textbf{FedProx}} & \multicolumn{2}{c}{\textbf{FedAdagrad}} & \multicolumn{2}{c}{\textbf{FedAdam}} & \multicolumn{2}{c}{\textbf{FedYogi}} \\
     \cmidrule(lr){2-3} \cmidrule(lr){4-5} \cmidrule(lr){6-7} \cmidrule(lr){8-9} \cmidrule(lr){10-11} \cmidrule(lr){12-13}
     & MU$\uparrow$ & FTR$\uparrow$ & MU$\uparrow$ & FTR$\uparrow$ & MU$\uparrow$ & FTR$\uparrow$ & MU$\uparrow$ & FTR$\uparrow$ & MU$\uparrow$ & FTR$\uparrow$ & MU$\uparrow$ & FTR$\uparrow$\\
    \toprule
    \rowcolor{cyan!5}
    \multicolumn{13}{c}{\textbf{Meta Llama-3.2-1B-Instruct with LoRA}} \\
    \midrule
    \rowcolor{gray!10} \textbf{Finetune} & 0.50 & 0.49 & 0.48 & 0.41 & 0.50 & 0.49 & 0.45 & 0.63 &0.48&0.53&0.48&0.54 \\
    \textbf{GradAscent} & \cellcolor{cyan!20}{\textbf{0.45}} & 0.58 & 1.3e-5 & \underline{0.66} & 0.43 & 0.58 & 0.33 & \underline{0.66} & 0.43 & 0.65 & 0.43 & 0.65 \\
    \textbf{GradDiff} & \cellcolor{cyan!20}{\textbf{0.45}} & 0.57 & 0.00086 & 0.67 & 0.43 & 0.56 & 0.42 & \underline{0.71} & \cellcolor{cyan!20}{\textbf{0.45}} & 0.62 & \cellcolor{cyan!20}{\textbf{0.45}} & 0.62 \\
    \textbf{NPO} & 0.46 & 0.58 & 9e-5 & 0.68 & 0.45 & 0.58 & 0.45 & \underline{0.69} & \cellcolor{cyan!20}{\textbf{0.47}} & 0.63 & \cellcolor{cyan!20}{\textbf{0.47}} & 0.63 \\
    \textbf{SimNPO} & 0.47 & 0.58 & 0.0017 & 0.65 & 0.45 & 0.58 & 0.49 & \underline{0.66} & \cellcolor{cyan!20}{\textbf{0.48}} & 0.63 & \cellcolor{cyan!20}{\textbf{0.48}} & 0.63 \\
    \textbf{Retrain} & \cellcolor{cyan!20}{\textbf{0.51}} & 0.63 & 0.49 & 0.57 & 0.50 & 0.63 & 0.46 & 0.69 & 0.29 & \underline{0.72} & 0.29 & \underline{0.72} \\
    \rowcolor{cyan!5}
    \midrule
    \multicolumn{13}{c}{\textbf{Meta Llama-3.2-3B-Instruct with LoRA}} \\
    \midrule
    \rowcolor{gray!10} \textbf{Finetune} & \cellcolor{cyan!20}{\textbf{0.59}} & 0.47 & 0.56 & 0.42 & 0.58 & 0.48 & 0.53 & \underline{0.60} & 0.50 & 0.56 & 0.50 & 0.56 \\
    \textbf{GradAscent} & \cellcolor{cyan!20}{\textbf{0.53}} & 0.60 & 0.00043 & \underline{0.79} & 0.46 & 0.58 & 0.43 & 0.70 & 0.49 & 0.63 & 0.50 & 0.63 \\
    \textbf{GradDiff} & 0.54 & 0.55 & 0 & \underline{0.69} & 0.50 & 0.53 & 0.47 & 0.66 & \cellcolor{cyan!20}{\textbf{0.55}} & 0.61 & \cellcolor{cyan!20}{\textbf{0.55}} & 0.60 \\
    \textbf{NPO} & 0.52 & 0.56 & 0.0015 & \underline{0.71} & 0.50 & 0.57 & 0.50 & 0.68 & 0.55 & 0.60 & \cellcolor{cyan!20}{\textbf{0.56}} & 0.61 \\
    \textbf{SimNPO} & 0.53 & 0.57 & 0.014 & 0.64 & 0.51 & 0.57 & 0.53 & \underline{0.65} & \cellcolor{cyan!20}{\textbf{0.56}} & 0.62 & 0.55 & 0.57 \\
    \textbf{Retrain} & \cellcolor{cyan!20}{\textbf{0.58}} & 0.62 & 0.56 & 0.58 & 0.57 & 0.62 & 0.53 & \underline{0.66} & 0.51 & 0.62 & 0.51 & 0.62 \\
    \midrule
    \rowcolor{cyan!5}
    \multicolumn{13}{c}{\textbf{Meta Llama-3.1-8B-Instruct with LoRA}} \\
    \midrule
    \rowcolor{gray!10} \textbf{Finetune} & \cellcolor{cyan!20}{\textbf{0.64}} & 0.45 & 0.21 & 0.45 & 0.6 & 0.45 & 0.55 & \underline{0.52} & 0.53 & 0.44 & 0.54 & 0.44 \\
    \textbf{GradAscent} & \cellcolor{cyan!20}{\textbf{0.57}} & 0.58 & 0.02 & 0.46 & 0.37 & 0.55 & 0 & 9.2e-09 & 0.47 & \underline{0.62} & 0.47 & \underline{0.62} \\
    \textbf{GradDiff} & \cellcolor{cyan!20}{\textbf{0.57}} & 0.54 & 0.13 & 0.50 & 0.52 & 0.48 & 0.53 & 1.2e-07 & 0.51 & 0.56 & 0.51 & \underline{0.57} \\
    \textbf{NPO} & \cellcolor{cyan!20}{\textbf{0.56}} & 0.56 & 0.53 & 0.53 & 0.52 & 0.56 & 0.48 & \underline{0.61} & 0.53 & 0.58 & 0.53 & 0.58 \\
    \textbf{SimNPO} & \cellcolor{cyan!20}{\textbf{0.59}} & 0.55 & 0.55 & 0.52 & 0.56 & 0.52 & 0.53 & 0.52 & 0.55 & \underline{0.57} & 0.56 & \underline{0.57} \\
    \textbf{Retrain} & \cellcolor{cyan!20}{\textbf{0.64}} & 0.60 & 0.20 & 0.58 & 0.56 & \underline{0.61} & 0.52 & 0.66 & 0.54 & 0.58 & 0.54 & 0.58 \\
    \bottomrule
  \end{tabular}
  \caption{Performance comparison of federated learning and unlearning algorithms on the \textbf{TOFU} dataset using \textbf{Llama-3.1-8B, Llama-3.2-1B and 3B} models, evaluated on metrics MU (Model Utility) and FTR (Forget Truth Ratio) with \textbf{Split95} strategies. Scores in \cellcolor{cyan!20}{\textbf{Bold}} indicate the optimal MU in different FL methods, while scores \underline{underlined} indicate the optimal FTR in different FL methods.}
  \label{suptable1}
\end{table*}

\begin{table*}[htp]
  \centering
  \begin{tabular}{ccccccccccccc}
  \toprule
    \multirow{3}[3]{*}{\textbf{Algorithms}} & \multicolumn{6}{c}{\textbf{Weighted Averaging-Based FL}} & \multicolumn{6}{c}{\textbf{Adaptive Optimization FL}} \\
     \cmidrule(lr){2-7} \cmidrule(lr){8-13}
     & \multicolumn{2}{c}{\textbf{FedAvg}} & \multicolumn{2}{c}{\textbf{FedAvgM}} & \multicolumn{2}{c}{\textbf{FedProx}} & \multicolumn{2}{c}{\textbf{FedAdagrad}} & \multicolumn{2}{c}{\textbf{FedAdam}} & \multicolumn{2}{c}{\textbf{FedYogi}} \\
     \cmidrule(lr){2-3} \cmidrule(lr){4-5} \cmidrule(lr){6-7} \cmidrule(lr){8-9} \cmidrule(lr){10-11} \cmidrule(lr){12-13}
     & MU$\uparrow$ & FTR$\uparrow$ & MU$\uparrow$ & FTR$\uparrow$ & MU$\uparrow$ & FTR$\uparrow$ & MU$\uparrow$ & FTR$\uparrow$ & MU$\uparrow$ & FTR$\uparrow$ & MU$\uparrow$ & FTR$\uparrow$\\
    \toprule
        \rowcolor{cyan!5}
    \multicolumn{13}{c}{\textbf{Meta Llama-3.2-1B-Instruct with LoRA}} \\
    \midrule
    \rowcolor{gray!10} \textbf{Finetune} & 0.50 & 0.49 & 0.48 & 0.45 & 0.50 & 0.49 & 0.45 & 0.62 & 0.45 & 0.60 & 0.45 & 0.59 \\
    \textbf{GradAscent} & \cellcolor{cyan!20}{\textbf{0.46}} & 0.61 & 0 & 0.050 & 0.43 & 0.64 & 0.40 &  \underline{0.72} & 0.44 & 0.65 & \cellcolor{cyan!20}{\textbf{0.46}} & 0.66 \\
    \textbf{GradDiff} & \cellcolor{cyan!20}{\textbf{0.46}} & 0.63 & 6.5e-5 &  \underline{0.70} & 0.44 & 0.60 & 0.42 &  \underline{0.70} & 0.44 & 0.66 & 0.44 & 0.67 \\
    \textbf{NPO} & \cellcolor{cyan!20}{\textbf{0.46}} & 0.62 & 2.9e-5 & 0.71 & 0.44 & 0.63 & 0.41 & \underline{0.74} & 0.45 & 0.68 & 0.45 & 0.68 \\
    \textbf{SimNPO} & \cellcolor{cyan!20}{\textbf{0.46}} & 0.65 & 0.00018 & 0.69 & 0.43 & 0.66 & 0.42 & \underline{0.74} & \cellcolor{cyan!20}{\textbf{0.46}} & 0.69 & \cellcolor{cyan!20}{\textbf{0.46}} & 0.70 \\
    \textbf{Retrain} & \cellcolor{cyan!20}{\textbf{0.51}} & 0.65 & 0.47 & 0.62 & \cellcolor{cyan!20}{\textbf{0.51}} & 0.64 & 0.46 &  \underline{0.67} & 0.46 & 0.66 & 0.46 & 0.66 \\
    \midrule
    \rowcolor{cyan!5}
    \multicolumn{13}{c}{\textbf{Meta Llama-3.2-3B-Instruct with LoRA}} \\
    \midrule
    \rowcolor{gray!10} \textbf{Finetune} & 0.59 & 0.49 & 0.56 & 0.48 & 0.58 & 0.51 & 0.53 & 0.61 & 0.50 & 0.57 & 0.50 & 0.57 \\
    \textbf{GradAscent} & \cellcolor{cyan!20}{\textbf{0.52}} & 0.59 & 0.00015 & \underline{0.79} & 0.48 & 0.62 & 0.45 & 0.73 & \cellcolor{cyan!20}{\textbf{0.52}} & 0.66 & 0.51 & 0.66 \\
    \textbf{GradDiff} & \cellcolor{cyan!20}{\textbf{0.52}} & 0.59 & 0.00062 &  \underline{0.77} & 0.49 & 0.59 & 0.47 & 0.71 & 0.51 & 0.61 & 0.51 &0.61\\
    \textbf{NPO} & \cellcolor{cyan!20}{\textbf{0.50}} & 0.62 & 0.00032 & \underline{0.79} & 0.47 & 0.60 & 0.45 & 0.73 & \cellcolor{cyan!20}{\textbf{0.50}} & 0.63 & \cellcolor{cyan!20}{\textbf{0.50}} &0.63\\
    \textbf{SimNPO} & \cellcolor{cyan!20}{\textbf{0.51}} & 0.61 & 0.0013 &  \underline{0.77} & 0.48 & 0.62 & 0.47 & 0.73 & 0.50 & 0.63 & \cellcolor{cyan!20}{\textbf{0.51}} & 0.65\\
    \textbf{Retrain} & \cellcolor{cyan!20}{\textbf{0.59}} & 0.64 & 0.56 & 0.64 & 0.57 & 0.65 & 0.53 &  \underline{0.66} & 0.50 & 0.63 & 0.50 & 0.63\\
    \midrule
    \rowcolor{cyan!5}
    \multicolumn{13}{c}{\textbf{Meta Llama-3.1-8B-Instruct with LoRA}} \\
    \midrule
    \rowcolor{gray!10} \textbf{Finetune} & \cellcolor{cyan!20}{\textbf{0.64}} & 0.49 & 0.21 & 0.51 & 0.6 & 0.48 & 0.55 & \underline{0.54} & 0.53 & 0.49 & 0.54 & 0.49 \\
    \textbf{GradAscent} & \cellcolor{cyan!20}{\textbf{0.57}} & \underline{0.63} & 0 & 0.44 & 0.37 & 0.53 & 0 & 0.43 & 0.47 & \underline{0.63} & 0.47 & \underline{0.63} \\
    \textbf{GradDiff} & \cellcolor{cyan!20}{\textbf{0.58}} & \underline{0.64} & 0.32 & 0.52 & 0.44 & 0.4 & 0.42 & 0.021 & 0.49 & 0.63 & 0.49 & 0.63 \\
    \textbf{NPO} & \cellcolor{cyan!20}{\textbf{0.58}} & 0.6 & 0.53 & 0.53 & 0.46 & 0.58 & 0.43 & \underline{0.7} & 0.48 & 0.63 & 0.48 & 0.63 \\
    \textbf{SimNPO} & \cellcolor{cyan!20}{\textbf{0.57}} & 0.64 & 0.55 & 0.58 & 0.46 & 0.57 & 0.42 & 0.64 & 0.48 & \underline{0.66} & 0.48 & \underline{0.66} \\
    \textbf{Retrain} & \cellcolor{cyan!20}{\textbf{0.62}} & 0.66 & 0.17 & 0.67 & 0.6 & 0.63 & 0.53 & \underline{0.69} & 0.54 & 0.66 & 0.54 & 0.66 \\
    \bottomrule
  \end{tabular}
  \caption{Performance comparison of federated learning and unlearning algorithms on the \textbf{TOFU} dataset using \textbf{Llama-3.1-8B, Llama-3.2-1B and 3B} models, evaluated on metrics MU (Model Utility) and FTR (Forget Truth Ratio) with \textbf{Split99} strategies. Scores in \cellcolor{cyan!20}{\textbf{Bold}} indicate the optimal MU in different FL methods, while scores \underline{underlined} indicate the optimal FTR in different FL methods.}
  \label{suptable_8B}
\end{table*}

\begin{table*}[htp]
\centering
\resizebox{\textwidth}{!}{
\begin{tabular}{ccccccccccccccccccc}
\toprule

\multirow{3}[3]{*}{\textbf{Algorithms}} & \multicolumn{9}{c}{\textbf{Weighted Averaging-Based FL}} & \multicolumn{9}{c}{\textbf{Adaptive Optimization FL}} \\
\cmidrule(lr){2-10} \cmidrule(lr){11-19} & \multicolumn{3}{c}{\textbf{FedAvg}} & \multicolumn{3}{c}{\textbf{FedAvgM}} & \multicolumn{3}{c}{\textbf{FedProx}} & \multicolumn{3}{c}{\textbf{FedAdagrad}} & \multicolumn{3}{c}{\textbf{FedAdam}} & \multicolumn{3}{c}{\textbf{FedYogi}} \\ 
\cmidrule(lr){2-4} \cmidrule(lr){5-7} \cmidrule(lr){8-10} \cmidrule(lr){11-13} \cmidrule(lr){14-16} \cmidrule(lr){17-19}
& \multicolumn{1}{c}{NVM} & \multicolumn{1}{c}{NKM} & \multicolumn{1}{c}{UP} & \multicolumn{1}{c}{NVM} & \multicolumn{1}{c}{NKM} & \multicolumn{1}{c}{UP} & \multicolumn{1}{c}{NVM} & \multicolumn{1}{c}{NKM} & \multicolumn{1}{c}{UP} & \multicolumn{1}{c}{NVM} & \multicolumn{1}{c}{NKM} & \multicolumn{1}{c}{UP} & \multicolumn{1}{c}{NVM} & \multicolumn{1}{c}{NKM} & \multicolumn{1}{c}{UP} & \multicolumn{1}{c}{NVM} & \multicolumn{1}{c}{NKM} & \multicolumn{1}{c}{UP} \\ 
\midrule
\textbf{Finetune} & 0.17 & \underline{0.13} & 0.18 & \underline{0.16} & 0.15 & 0.14 & \underline{0.16} & 0.15 & \cellcolor{cyan!20}{\textbf{0.21}} & \underline{0.16} & \underline{0.13} & 0.17 & 0.17 & 0.15 & 0.18 & 0.17 & 0.14 & 0.17 \\
\textbf{GradAscent} & 0.62 & 0.4 & \cellcolor{cyan!20}{\textbf{0.63}} & 0.48 & 0.3 & 0.47 & 0.31 & 0.38 & 0.61 & \underline{0.045} & \underline{0.014} & 0.016 & 0.39 & 0.45 & 0.5 & 0.51 & 0.39 & 0.43 \\
\textbf{GradDiff} & 0.14 & 0.41 & \cellcolor{cyan!20}{\textbf{0.64}} & 0.32 & \underline{0.35} & 0.63 & \underline{0} & 0.37 & 0.56 & 0.0036 & 0.38 & 0.58 & 0.11 & 0.38 & 0.57 & 0.11 & 0.43 & 0.63 \\
\textbf{NPO} & 0.99 & \underline{0.39} & 0.65 & 0.99 & 0.43 & 0.63 & 0.99 & 0.4 & 0.62 & 0.97 & 0.41 & 0.62 & 1.0 & 0.4 & 0.65 & \underline{0.94} & 0.43 & \cellcolor{cyan!20}{\textbf{0.66}} \\
\textbf{SimNPO} & 0.23 & 0.33 & 0.61 & 0.47 & 0.39 & 0.63 & \underline{0.047} & \underline{0.24} & 0.57 & 0.073 & 0.25 & 0.58 & 0.19 & 0.34 & 0.6 & 0.14 & 0.32 & \cellcolor{cyan!20}{\textbf{0.64}} \\
\textbf{Retrain} & \underline{0.16} & 0.15 & 0.18 & \underline{0.16} & \underline{0.12} & 0.13 & \underline{0.16} & 0.15 & 0.19 & \underline{0.16} & 0.15 & 0.17 & \underline{0.16} & 0.16 & \cellcolor{cyan!20}{\textbf{0.19}} & \underline{0.16} & 0.15 & 0.18 \\
\bottomrule
\end{tabular}
}
\caption{Performance comparison of federated learning algorithms on the \textbf{MUSE Books} set using \textbf{Llama-2-7B model}, evaluated on metrics NVM (No Verbatim Mem$\downarrow$), NKM (No Knowledge Mem$\downarrow$), and UP (Utility Preserved$\uparrow$). Scores in \cellcolor{cyan!20}{\textbf{Bold}} indicate the optimal UP in different FL methods, while \underline{underlined} indicate the optimal NVM and NKM in different FL methods.}
\label{suptable3}
\end{table*}

\clearpage

\begin{figure*}[t]
\begin{minipage}{\linewidth}
\begin{cmt}[]{Forget Set}
\textbf{Input}: Which city and country was Basil Mahfouz Al-Kuwaiti born? \\

\textbf{Ground Truth}: Basil Mahfouz Al-Kuwaiti was born in Kuwait City, Kuwait. \\

\textbf{Generation}: Basil Mahfouz Al-Kuwaiti was born in Al-Qadisiyah, Kuwait.
\end{cmt}
\end{minipage}

\medskip
\medskip
\medskip

\begin{minipage}{\linewidth}
\begin{cmt}[]{Retain Set}
\textbf{Input}: Jaime Vasquez earned any awards for his controversial works? \\
\textbf{Ground Truth}: Jaime Vasquez was bestowed with the prestigious Edgar Allan Poe Award for Best Fact Crime, which is an illustrious accolade in the domain of crime fiction and non-fiction. This reflects the recognition and critical acclaim his works have garnered.\\
\textbf{Generation}: Yes, Jaime Vasquez has received the prestigious Edgar Allan Poe Award for Best Fact Crime for his work \"The Devil's Handshake\".
\end{cmt}
\end{minipage}

\medskip
\medskip
\medskip

\begin{minipage}{\linewidth}
\begin{cmt}[]{Real Authors Set}
\textbf{Input}: Who penned 'The Hobbit' and 'The Lord of the Rings' series? \\

\textbf{Ground Truth}: Tolkien, \\

\textbf{Generation}: 'The Hobbit' and 'The Lord of the Rings' series were penned by J.R.R. Tolkien.
\end{cmt}
\end{minipage}

\medskip
\medskip
\medskip

\begin{minipage}{\linewidth}
\begin{cmt}[]{World Facts Set}
\textbf{Input}: Which country is known as the Land of the Rising Sun? \\

\textbf{Ground Truth}: Japan \\

\textbf{Generation}: Japan, known as the Land of the Rising Sun.
\end{cmt}
\end{minipage}

\caption{Comparison of model outputs for the \textbf{Forget}, \textbf{Retain}, \textbf{Real Authors}, and \textbf{World Facts} sets using \textbf{FedYogi} and \textbf{NPO} algorithms.}
\label{fig:tofu_qa_examples_2x2}
\end{figure*}

\clearpage